\documentclass{ws-ijait}

\usepackage{times}
\usepackage[small,compact]{titlesec}

\usepackage{amsmath}
\usepackage{amsmath}
\usepackage{amsfonts}
\usepackage{amssymb}
\usepackage{type1cm}
\usepackage{bbm}

\usepackage{graphicx}
\usepackage{epstopdf}
\DeclareGraphicsExtensions{.pdf,.eps,.png,.jpg,.tif,.tiff}
\RequirePackage{epsfig}
\usepackage[caption=false]{subfig}
\graphicspath{{./}{img/}}
\usepackage{fixltx2e}

\usepackage{booktabs}
\usepackage{chngpage}
\usepackage{rotating}
\usepackage{multirow}
\usepackage{array}

\RequirePackage{color}
\RequirePackage{xspace}
\RequirePackage{eso-pic}
\usepackage{ucs}
\usepackage[utf8x]{inputenc}	
\usepackage{url}
\usepackage{hyperref}
\usepackage[svgnames]{xcolor}

\usepackage{algorithm}
\usepackage{algorithmic}

\usepackage{xspace}
\newcommand{\KMeans}{\mbox{K-Means}}%
\newcommand{\TDCKMeans}{\mbox{TDCK-Means}}%

\newcommand{\squishlist}{
 \begin{list}{$\bullet$}
  { \setlength{\itemsep}{0pt}
     \setlength{\parsep}{3pt}
     \setlength{\topsep}{3pt}
     \setlength{\partopsep}{0pt}
     \setlength{\leftmargin}{1.5em}
     \setlength{\labelwidth}{1em}
     \setlength{\labelsep}{0.5em} } }

\newcommand{\squishlisttwo}{
 \begin{list}{$\bullet$}
  { \setlength{\itemsep}{0pt}
    \setlength{\parsep}{0pt}
    \setlength{\topsep}{0pt}
    \setlength{\partopsep}{0pt}
    \setlength{\leftmargin}{1.5em}
    \setlength{\labelwidth}{1.5em}
    \setlength{\labelsep}{0.5em} } }

\newcommand{\squishend}{
  \end{list}  }

\hypersetup
{
pdftitle={How to Use Temporal-Driven Constrained Clustering to Detect Typical Evolutions},%
pdfauthor={M.A. Rizoiu, J. Velcin, S. Lallich},%
pdfkeywords={semi-supervised clustering; temporal clustering; temporal-aware dissimilarity measure; contiguity penalty function; temporal cluster graph structure.}%
}

\begin{document}

\markboth{Rizoiu et al.}
{How to Use Temporal-Driven Constrained Clustering to Detect Typical Evolutions}

%
\catchline{}{}{}{}{}
%

\title{ \uppercase{How to Use Temporal-Driven Constrained Clustering to Detect Typical Evolutions}
}

\author{\footnotesize MARIAN-ANDREI RIZOIU }

\address{ERIC Laboratory, University Lumi\`ere Lyon 2\\
5, avenue Pierre Mendès France \\
69676 Bron Cedex, France \\
Marian-Andrei@rizoiu.eu}

\author{\footnotesize JULIEN VELCIN }

\address{ERIC Laboratory, University Lumi\`ere Lyon 2\\
Julien.Velcin@univ-lyon2.fr}

\author{\footnotesize ST\'{E}PHANE LALLICH}

\address{ERIC Laboratory, University Lumi\`ere Lyon 2\\
Stephane.Lallich@univ-lyon2.fr}

\maketitle

\begin{history}
\received{(Day Month Year)}
\revised{(Day Month Year)}
\accepted{(Day Month Year)}
\end{history}

\begin{abstract}
In this paper, we propose a new time-aware dissimilarity measure that takes into account the temporal dimension.
Observations that are close in the description space, but distant in time are considered as dissimilar.
We also propose a method to enforce the segmentation contiguity, by introducing, in the objective function, a penalty term inspired from the Normal Distribution Function.
We combine the two propositions into a novel time-driven constrained clustering algorithm, called \textbf{\TDCKMeans{}}, which creates a partition of coherent clusters, both in the multidimensional space and in the temporal space.
This algorithm uses soft semi-supervised constraints, to encourage adjacent observations belonging to the same entity to be assigned to the same cluster.
We apply our algorithm to a Political Studies dataset in order to detect typical evolution phases.
We adapt the Shannon entropy in order to measure the entity contiguity, and we show that our proposition consistently improves temporal cohesion of clusters, without any significant loss in the multidimensional variance.
\end{abstract}


\section{Introduction}
\label{sec:introduction}

Researchers in Social Sciences and Humanities (like Political Studies) have always gathered data and compiled databases of knowledge.
This information often has a temporal component, the evolution of a certain number of entities is recorded over a period of time.
Each entity is described using multiple attributes, which form the multidimensional description space.
Therefore, an entry in such a database would be an observation, a triple $(entity, timestamp, description)$.
An observation $x_i = (\phi_l, t_m, x_i^d)$ signifies that the entity $\phi_l$ is described by the vector $x_i^d$ at the moment of time $t_m$.
We denote by $x_i^\phi$ the entity to which the observation $x_i$ is associated.
Similarly, $x_i^t$ is the timestamp associated with the observation $x_i$.
Each observation belongs to a single entity and, consequently, each entity is associated with multiple observations, for different moments of time.
Formally:
\begin{align*}
	\forall x_i \in \mathcal{D} &: \exists ! \, \phi_l \in \Phi \text{ so that } x_i^\phi = \phi_l \\
	\forall (\phi_l, t_m) \in \Phi \times \mathcal{T} &: \exists ! \, x_i=(x_i^\phi, x_i^t, x_i^d) \text{ so that } x_i^\phi = \phi_l \text{ and } x_i^t = t_m
\end{align*}
For example, a database studying the evolution of democratic states~\cite{ARM11} will store, for each country and each year, the value of multiple economical, social, political and financial indicators.
The countries are the entities, and the years are the timestamps.

Starting from such a database, one of the interests of Political Studies researchers is to detect typical evolution patterns. 
There is a double interest: 
a) obtaining a broader understanding of the phases that the entity collection went through over time (\textit{e.g.} detecting the periods of global political instability, of economic crisis, of wealthiness \textit{etc.});
b) constructing the trajectory of an entity through the different phases (\textit{e.g.} a country may have gone through a period of military dictatorship, followed by a period of wealthy democracy).
The criteria describing each phase are not known beforehand (which indicators announce a world economic crisis?) and may differ from one phase to another.

\begin{figure}[tb]
	\centering
	\subfloat[] {
		\includegraphics[width=0.45\textwidth]{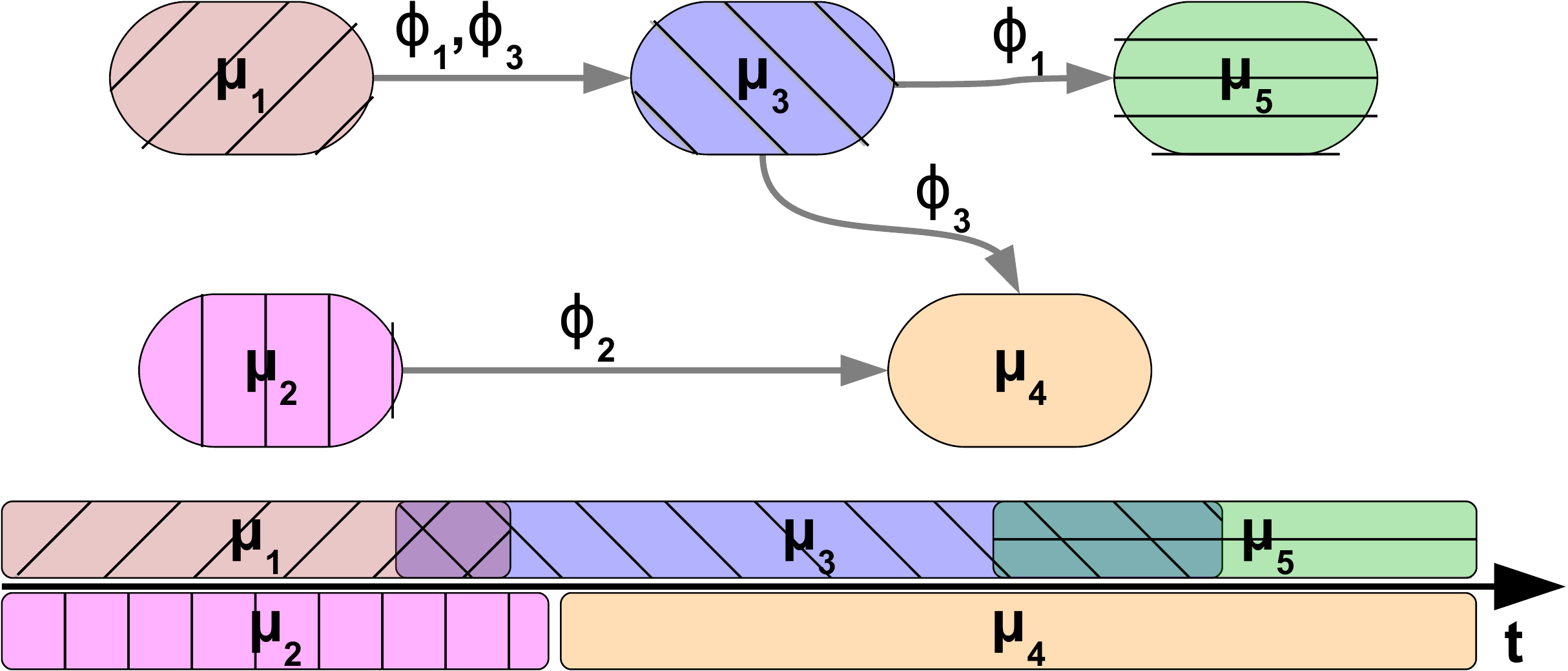}
		\label{subfig:cluster-structuring}
	}
	\hfill
	\subfloat[] {
		\includegraphics[width=0.45\textwidth]{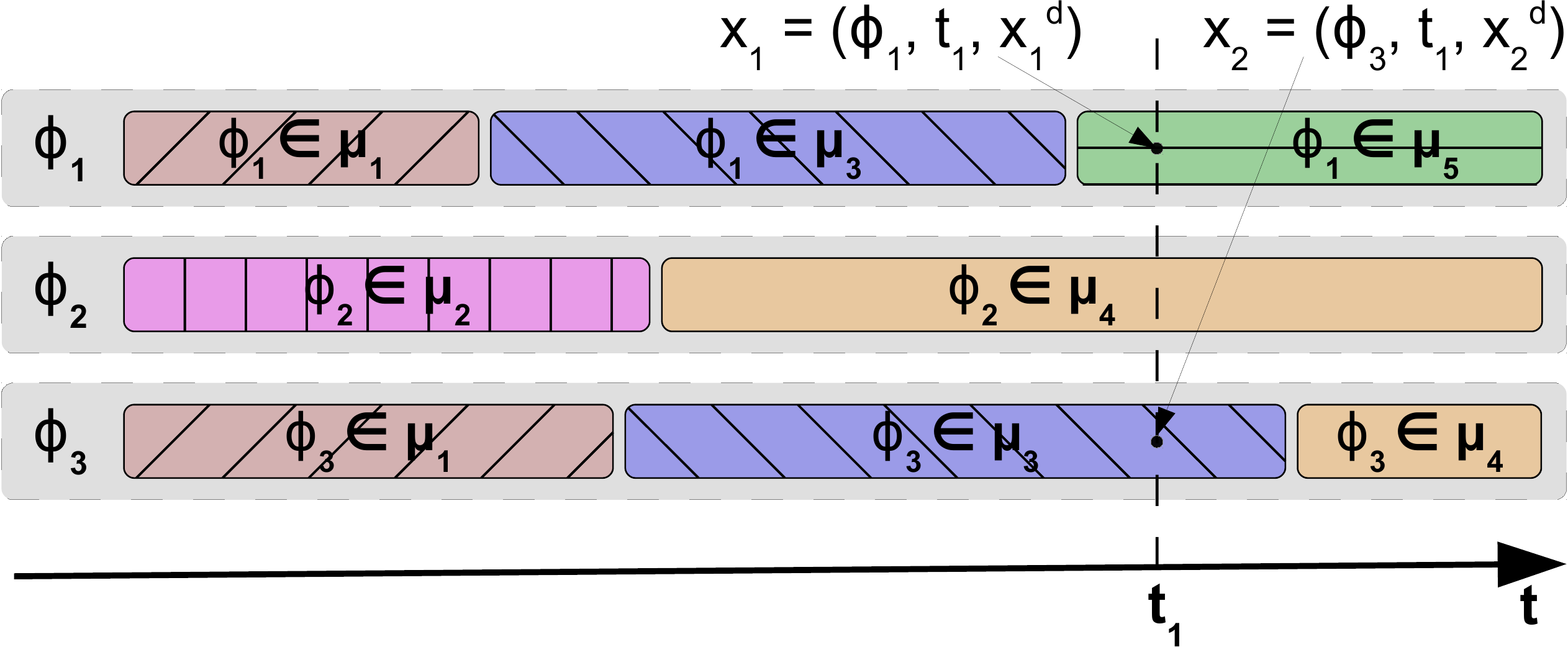}
		\label{subfig:entity-contiguity}
	}
	\caption{Desired output: (a) the evolution phases and the entity trajectories, (b) the observations of 3 entities contiguously partitioned into 5 clusters.}
	\label{fig:desired-output}
	\vspace{-0.2in}
\end{figure}

We address these issues by proposing a novel temporal-driven constrained clustering algorithm.
The proposed algorithm partitions the observations into clusters $\mu_j \in \mathcal{M}$, that are coherent both in the multidimensional description space and in the temporal space.
We consider that the obtained clusters can be used to represent the typical phases of the evolution of the entities through time. 
Figure~\ref{fig:desired-output} shows the desired result of our clustering algorithm.
Each of the three depicted entities ($\phi_1, \phi_2$ and $\phi_3$) is described at 10 moments of time ($t_m, m=1,2,...,10$).
The 30 observations of the dataset are partitioned into 5 clusters ($\mu_j, j=1,2,...,5$).
In Figure~\ref{subfig:cluster-structuring} we observe how clusters $\mu_j$ are organized in time.
Each of the clusters has a limited extent in time, and the time extents of clusters can overlap.
The temporal extent of a cluster is the minimal interval of time that contains all the timestamps of the observations in that cluster.
The entities navigate through clusters.
When an observation belonging to an entity is assigned to cluster $\mu_2$ and the anterior observation of the same entity is assigned in cluster $\mu_1$, then we consider that the entity has a transition from phase $\mu_1$ to phase $\mu_2$.
Figure~\ref{subfig:entity-contiguity} shows how the series of observations belonging to each entity are assigned to clusters, thus forming continuous segments.
This succession of segments is interpreted as the succession of phases through which the entity passes. 
For this succession to be meaningful, each entity should be assigned to a rather limited number of continuous segments.
Passing through too many phases reduces the comprehension.
Similarly, evolutions which are alternations between two phases (\textit{e.g.}, $\mu_1 \longrightarrow \mu_2 \longrightarrow \mu_1 \longrightarrow \mu_2 $) hinder the comprehension.

Based on these observations, we assume that the resulting partition must:
\squishlisttwo    
	\item \textbf{regroup observations having similar descriptions into the same cluster} (just as traditional clustering does).
	The clusters represent a certain type of evolution;
	\item \textbf{create temporally coherent clusters, with limited extent in time.} 
	In order for a cluster to be meaningful, it should regroup observations which are temporally close (be contiguous on the temporal dimension).
	If there are two different periods with similar evolutions (\textit{e.g.} two economical crises), it is preferable to have them regrouped separately, as they represent two distinct phases.
	Furthermore, while it is acceptable that some evolutions exist during the entire period, usually the resulted clusters should have a limited temporal extent;
	\item \textbf{segment, as contiguously as possible, the series of observations for each entity.} 
	The sequence of segments will be interpreted as the sequence of phases through which the entity passes.
\squishend    

In this paper, we propose a new time-aware dissimilarity measure that takes into account the temporal dimension.
Observations that are close in the description space, but distant in time are considered as dissimilar.
We also propose a method to enforce the segmentation contiguity, by introducing a penalty term based on the Normal Distribution Function.
We combine the two propositions into a novel time-driven constrained clustering algorithm, \textbf{\TDCKMeans{}}, which creates a partition of coherent clusters, both in the multidimensional space and in the temporal space.
This algorithm uses soft semi-supervised constraints to encourage adjacent observations belonging to the same entity to be assigned to the same cluster.
The proposed algorithm constructs the clusters that serve as evolution phases and segments the observations series for each entity.

The paper is organized as follows.
In Section~\ref{sec:related-work} we present some previous related works and, in Section~\ref{sec:TDCK-Means-proposal}, we introduce the temporal-aware dissimilarity function, the contiguity penalty, function  the \TDCKMeans{} algorithm and the graph structure induction method.
In Section~\ref{sec:xp}, we present the dataset that we use, the proposed evaluation measures and the obtained results.
Finally, in Section~\ref{sec:conclusion}, we draw the conclusion and plan some future extensions.

\section{Related work}
\label{sec:related-work}

Leveraging partial expert knowledge into clustering represents the domain of semi-supervised clustering.
The expert knowledge is under the form of either class labels, or pairwise constraints.
Pairwise constraints~\cite{WAG01} are either ``must-link'' (the observations must be placed in the same cluster) or ``cannot-link'' (the two observations cannot be placed in the same cluster).
Depending on the method in which supervision is introduced into clustering, \cite{GRI05} divides the semi-supervised clustering methods into two classes:
a) the similarity-adapting methods~\cite{BIL03,COH03,KLE02,XIN02}, which seek to learn new similarity measures in order to satisfy the constraints, and 
b) the search-based methods~\cite{BAS02,DEM99,WAG01} in which the clustering algorithm itself is modified.

The literature presents some examples of algorithms used to segment a series of observations into continuous chunks.
In \cite{LIN06}, the daily tasks of a user are detected by segmenting scenes from the recordings of his activities.
Semi-supervised must-link constraints are set between all pairs of observations, and a fixed penalty is inflicted when the following conditions are fulfilled simultaneously: the observations are not assigned to the same cluster and the time difference between their timestamps is less than a certain threshold.
A similar technique is used in \cite{TOR07}, where constraints are used to penalize non-smooth changes (over time) on the assigned clusters.
This segmenting technique is used to detect tasks performed during a day, based on video, on sound and on GPS information.
In~\cite{SAN01}, the objects appearing in an image sequence are detected by using a hierarchical descending clustering, that regroups pixels into large temporally coherent clusters.
This method seeks to maximize the cluster size, while guaranteeing intra-cluster temporal consistency.
All of these techniques consider only one series of observations (a single entity) and must be adapted for the case of multiple series.
The main problem of a threshold based penalty function is to set the value of the threshold, which is usually data-dependent.
Optimal matching is used in~\cite{WID09} to discover trajectory models, while studying  the de-standardization of typical life courses.

The temporal dimension of the data is also used in some other fields of Information Retrieval.
In \cite{TAL12}, constrained clustering is used to scope temporal relational facts in the knowledge bases, by exploiting temporal containment, alignment, succession, and mutual exclusion constraints among facts. 
In \cite{CHE09}, clustering is used to segment temporal observations into continuous chunks, as a preprocessing phase.
A graphical model is proposed in \cite{QAM06}, that uses a probabilistic model in which the timestamp is part of the observed variables, and the story is the hidden variable to be inferred.
But still, none of these approaches seek to create temporally coherent partitions of the data, mainly using the temporal dimension as a secondary information.

In the following sections, we propose a dissimilarity measure, a penalty function and a clustering algorithm in which the temporal dimension has a central role, and which address the limitations existing in the above presented work.

\section{Temporal-Driven Constrained Clustering}
\label{sec:TDCK-Means-proposal}

The observations $x_i \in \mathcal{X} $ that need to be structured can be written as triples $(entity, time, description)$: $x_i = (x_i^\phi, x_i^t, x_i^d)$.
$x_i^d \in \mathcal{D}$ is the vector in the multidimensional description space which describes the entity $x_i^\phi \in \Phi$ at the moment of time $x_i^t \in \mathcal{T}$.

Traditional clustering algorithms input a set of multidimensional vectors, which they regroup in such a way that observations inside a group resemble each other as much as possible, and resemble observations in other groups as little as possible.
\KMeans{}~\cite{MAC67} is a clustering algorithm based on iterative relocation, that partitions a dataset into $m$ clusters, locally minimizing the sum of distances between each data points $x_i$ and its assigned cluster centroids $\mu_j \in \mathcal{M}$.
At each iteration, the objective function \begin{equation*}
	{I} = \Sigma_{\mu_j \in \mathcal{M}}\Sigma_{x_i \in \mathcal{C}_j} || x_i^d - \mu_j^d ||^2
\end{equation*}
is minimized until it reaches a local optimum.

Such a system is appropriate for constructing partitions based solely on $x_i^d$, the description in the multidimensional space.
It does not take into account the temporal order of the observations, nor the structure of the dataset, the fact that observations belong to entities.
We extend to the temporal case by adding to the centroids a temporal dimension $\mu_j^t$, described in the same temporal space $\mathcal{T}$ as the observations.
Just like its multidimensional description vector $\mu_j^d$, the temporal component does not necessary need to exist in the temporal set of the observation.
It is an abstraction of the temporal information in the group, serving as a cluster timestamp.
Therefore, a centroid $\mu_j$ will be the couple $(\mu_j^t, \mu_j^d)$.

We propose to adapt the \KMeans{} algorithm to the temporal case by adapting the Euclidean distance, normally used to measure the distance between an element and its centroid.
This novel temporal-aware dissimilarity measure takes into account both the distance in the multidimensional space and in the temporal space.
In order to ensure the temporal contiguity of observations for the entities, we add a penalty whenever two observations that belong to the same entity are assigned to different clusters. 
The penalty depends on the time difference between the two: the lower the difference, the higher the penalty.
We integrate both into the \textbf{Temporal-Driven Constrained \KMeans{}} (\textbf{\TDCKMeans{}}), which is a temporal extension of \KMeans{}.
\TDCKMeans{} searches to minimize the following objective function:
\begin{equation} \label{eq:obj-function}
	\mathcal{J} = \sum_{\mu_j \in \mathcal{M}}\sum_{x_i \in \mathcal{C}_j} \left( || x_i - \mu_j||_{TA} + \sum_{\substack{x_k \not\in \mathcal{C}_j\\x_k^\phi = x_i^\phi}} w(x_i, x_k) \right)
\end{equation}
where $||\bullet||_{TA}$ is our temporal-aware (TA) dissimilarity measure (detailed in the next section), $w(x_i,x_j)$ is the cost function that determines the penalty of clustering adjacent observations of the same entity into different clusters, and $\mathcal{C}_j$ is the set of observations in cluster $j$.

\subsection{The temporal-aware dissimilarity measure}
\label{subsec:proposal-temporal-dissimilarity-measure}

The proposed temporal-aware dissimilarity measure \mbox{$|| x_i - x_j||_{TA}$} combines the Euclidean distance in the multidimensional space $\mathcal{D}$ and the distance between the timestamps.
We propose to use the following formula:
\begin{equation} \label{eq:temp-distance}
	|| x_i - x_j||_{TA} = 1 - \left(1 - \frac{||x_i^d - x_j^d||^2}{\Delta x_{max}^2}\right)\left(1 - \frac{||x_i^t - x_j^t||^2}{\Delta t_{max}^{2}}\right) 
\end{equation}
where $||\bullet||$ is the classical $L^2$ norm and $\Delta x_{max}$ and $\Delta t_{max}$ are the diameters of $\mathcal{D}$, and $\mathcal{T}$ respectively (the largest distance encountered between two observations in the multidimensional description space and, respectively, in the temporal space).
The following properties are immediate:
\squishlisttwo  
	\item $ || x_i - x_j||_{TA} \in [0,1], \forall x_i, x_j \in \mathcal{X} $
	\item $ || x_i - x_j||_{TA} = 0 \Leftrightarrow x_i^d = x_j^d$ and $x_i^t = x_j^t$
	\item $ || x_i - x_j||_{TA} = 1(maximum) \Leftrightarrow ||x_i^d - x_j^d|| = \Delta x_{max}$ or $||x_i^t - x_j^t|| = \Delta t_{max}$
\squishend   

\begin{figure}[!t]
	\centering
	\includegraphics[width=0.45\textwidth]{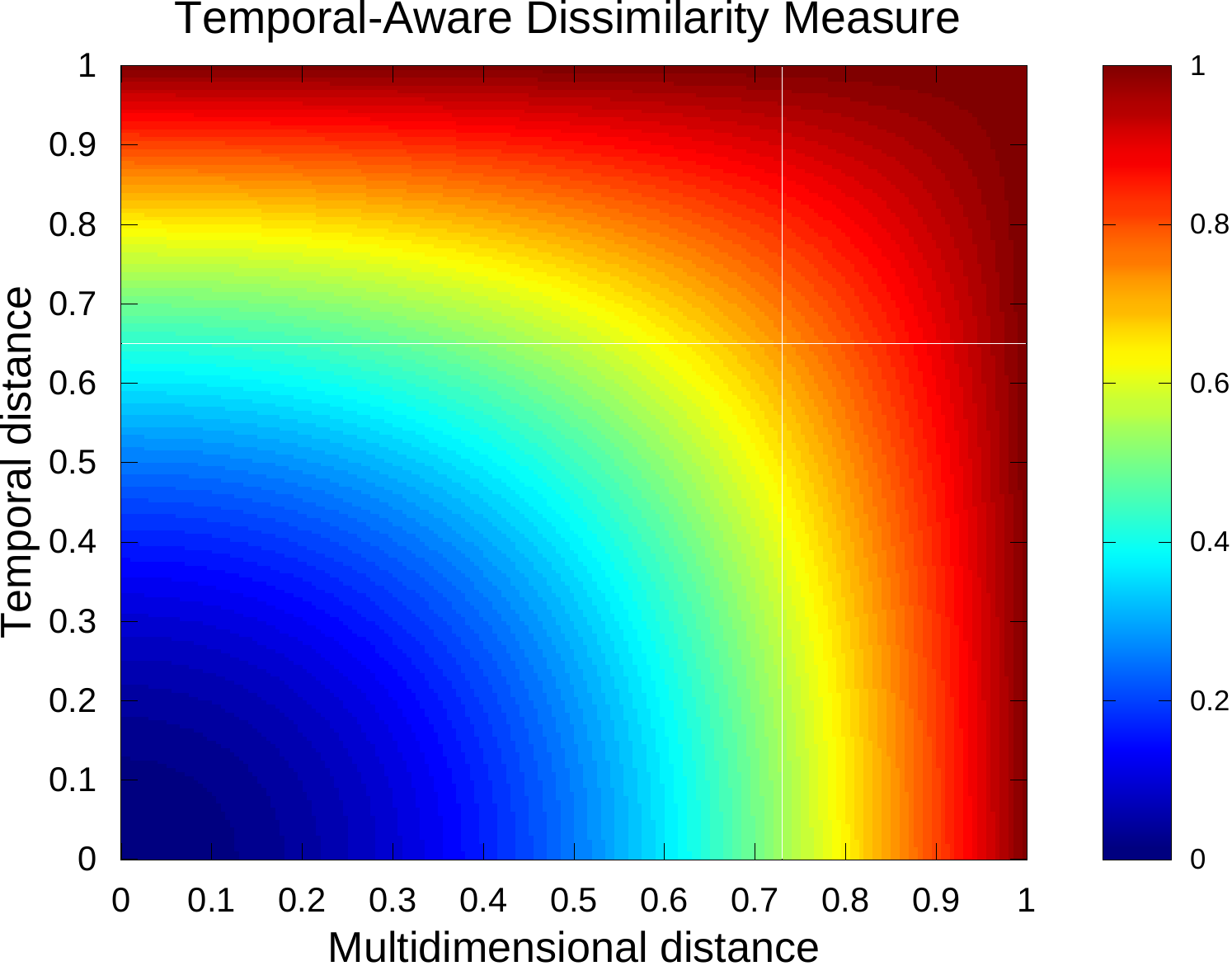}
	\caption{Color map of the temporal-aware dissimilarity measure as a function of the multidimensional component and the temporal component.}
	\label{fig:temporal-measure-colormap}
	\vspace{-0.2in}
\end{figure}

Figure~\ref{fig:temporal-measure-colormap} plots the temporal-aware dissimilarity measure as a color map, depending on the multidimensional component and the temporal component.
The horizontal axis represents the normalized multidimensional distance ($\frac{||x_i^d - x_j^d||^2}{\Delta x_{max}^2}$).
The vertical axis represents the normalized temporal distance ($\frac{||x_i^t - x_j^t||^2}{\Delta t_{max}^{2}}$).
The blue color shows a temporal-aware measure close to the minimum and the red color represents the maximum.
The dissimilarity measure is zero if and only if the two observations have equal timestamps and equal multidimensional description vectors.
Still, it suffices for only one of the components (temporal, multidimensional) to attend the maximum value for the measure to reach its maximum.
The measure behaves similar to a MAX operator, always choosing a value closer to the maximum of the two components.
The formula for the temporal-aware dissimilarity measure was chosen so that any algorithm that seeks to minimize an objective function based on this measure, will need to minimize both its components.
This makes it suitable for algorithms that search to minimize both the multidimensional and the temporal variance in clusters.

Both components that intervene in the measure follow a function like $ 1 - \epsilon^2, \epsilon \in [0,1]$.
This function provides a good compromise: it is tolerant for small values of $\epsilon$ (small time difference, small multidimensional distance), but decreases rapidly when $\epsilon$ augments.
The temporal-aware dissimilarity measure is an extension of the Euclidean function.
If the timestamps are unknown and set to be all equal, the temporal component is canceled and the temporal-aware dissimilarity measure becomes a normalized Euclidean distance.
In Section~\ref{subsec:quantitative-evaluation}, we evaluate the behavior of the proposed dissimilarity function.
We will call \textbf{Temporal-Driven \KMeans{}} the algorithm that is based on the \KMeans{}' iterative structure and uses the temporal-aware dissimilarity measure to asses similarity between observations.
Note that \textbf{Temporal-Driven \KMeans{}}, relative to \TDCKMeans{}, has no contiguous segmentation penalty function (the contiguous segmentation penalty function is detailed in the next section).

\subsection{The contiguity penalty function}
\label{subsec:proposal-penalty-function}

\begin{figure}[!t]
	\centering
	\includegraphics[width=0.45\textwidth]{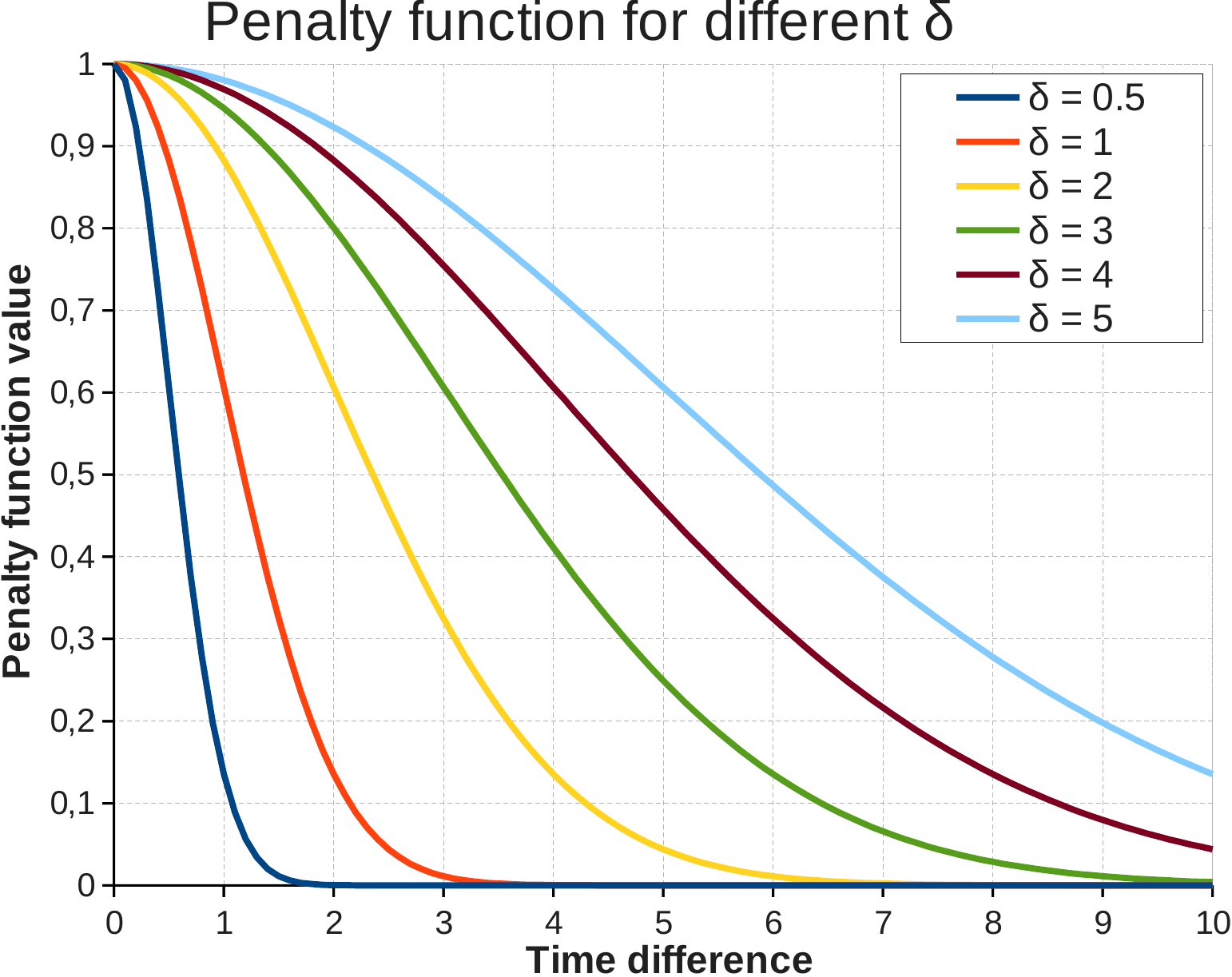}
	\caption{Penalty function vs. time difference for multiple $\delta$. $(\beta = 1)$}
	\label{fig:penalty-delta}
	\vspace{-0.2in}
\end{figure}

The penalty function encourages temporally adjacent observations of the same entity to  be assigned to the same cluster.
We use the notion of \textit{soft pair-wise constraints} from semi-supervised clustering.
A ``must-link'' soft constraint is added between all pairs of observations belonging to the same entity.
The clustering is allowed to break the constraints, while inflicting a penalty for each of these violations.
The penalty is more severe if the observations are closer in time.
The function is defined as:
\begin{equation} \label{eq:penalty-function}
	w(x_i, x_k) = \beta * e^{-\frac{1}{2} \left( \frac{||x_i^t - x_k^t||}{\delta} \right)^2} \mathbbm{1} \left[x_i^\phi = x_k^\phi \right] 
\end{equation}
where $\beta$ is a scaling factor and, at the same time, the maximum value taken by the penalty function;
$\delta$ is a parameter which controls the width of the function.
$\beta$ is dataset dependent and can be set as a percentage of the average distance between observations.
$\mathbbm{1} \left[ statement \right]$ is a function that returns $1$ if $statement$ is true and $0$ otherwise.

The function resembles to the positive side of the Normal Distribution function, centered in zero.
The function has a particular shape, as represented in Figure~\ref{fig:penalty-delta}.
For small time differences, it descends very slowly, thus inflicting a high penalty for breaking a constraint.
As the time difference increases, the penalty decreases rapidly, converging towards zero.
When $\delta$ is small, the functions value descends very quickly with the time difference.
The function produces penalties only if the constraint is broken for adjacent observation.
For high values of $\delta$, breaking constraints for distant observations cause high penalties, therefore creating segmentations with large segments.
Figure~\ref{fig:penalty-delta} shows the evolution of the penalty function with the time difference between two observations, for multiple values of $\delta$ and for $\beta=1$.

An advantage of the proposed function is that it requires no time discretization or setting a fixed window width, as proposed in \cite{LIN06}.
The $\delta$ parameter permits the fine tuning of the penalty function.
In Section~\ref{subsec:quantitative-evaluation}, we evaluate \textbf{Constrained \KMeans{}}, which is an extension of \KMeans{}, to which we add the proposed contiguity penalty function (but which does not take into account the temporal dimension when measuring the distance between observations).
The influence of both $\beta$ and $\delta$ will be studied in Section~\ref{subsec:parameters-beta-delta}.

\subsection{The \TDCKMeans{} algorithm}
\label{subsec:proposal-tdck-means}

The time dependent distance $|| x_i - \mu_j ||_{TA}$ encourages the decrease of both the temporal and multidimensional variance of clusters;
meanwhile the penalty function $w(x_i,x_k)$ favors the adjacent observations belonging to the same entity to be assigned to the same cluster.
The rest of the \TDCKMeans{} algorithm is similar to the \KMeans{} algorithm.
It seeks to minimize $\mathcal{J}$ by iterating an assignment phase and a centroid update phase until the partition does not change between two iterations.
The outline of the algorithm is given in Algorithm~\ref{algo:proposed-algo}.

The \textbf{choose\_random} function chooses randomly, for each centroid $\mu_j$, an observation $x_i$ and sets $\mu_j = (x_i^t, x_i^d)$.
In the assignment phase, for every observation $x_i$, the \textbf{best\_cluster} function chooses a cluster $\mathcal{C}_j$ so that the temporal-aware dissimilarity measure from $x_i$ to the cluster's centroid $\mu_j$, added to the cost of penalties possibly incurred by this cluster assignment, is minimized.
It resumes to solving the following equation:
\begin{equation*}
\mathbf{best\_cluster}(i) = \underset{j = 1,2,...,m}{argmin} \left( || x_i - \mu_j^{(iter-1)}\,\,||_{TA}^2 + \sum_{\substack{x_k \not\in \mathcal{C}_j^{(iter-1)}}}^{x_k^\phi = x_i^\phi} w(x_i, x_k) \right)
\end{equation*}
This guaranties that the contribution of $x_i$ to the value of $\mathcal{J}$ diminishes or stays constant.
Overall, this assures that $\mathcal{J}$ diminishes in the assignment phase (or stays constant).

\begin{algorithm}[ht]                     
\caption{Outline of the \TDCKMeans{} algorithm.}         
\label{algo:proposed-algo} 

\begin{algorithmic} 

\REQUIRE $ x_i \in \mathcal{X} $ - observations to cluster;
\REQUIRE $ m $ - number of requested clusters;
\ENSURE $\mathcal{C}_j, j = 1,2,...,m$ - $m$ clusters;
\ENSURE $\mu_j, j = 1,2,...,m$ - centroids for each cluster; 

\FOR {$j = 1,2,..,m$}
	\STATE $ \mu_j \gets $ \textbf{choose\_random}($\mathcal{X}$)
\ENDFOR
\STATE $ iter \gets 0$
\STATE $\mathcal{M}^{(iter)} \gets \emptyset$ \hspace{5mm} \textit{//set of centroids}
\STATE $\mathcal{P}^{(iter)} \gets \emptyset$ \hspace{5mm} \textit{//set of clusters}

\REPEAT	
	\STATE $iter \gets iter + 1$
	\FOR {$j = 1,2,...,m $}
		\STATE $\mathcal{C}_j^{(iter)} \gets \emptyset$
	\ENDFOR
	
	\STATE \textit{// assignment phase}
	\FOR { $x_i \in \mathcal{X}$ }
		\STATE $\mathcal{C}_j^{(iter)} = \mathcal{C}_j^{(iter)} \cup x_i | $ where $ j = $ \textbf{best\_cluster}($\mathcal{X}$, $\mathcal{M}^{(iter-1)}$, $\mathcal{P}^{(iter-1)}$)
	\ENDFOR
	
	\STATE \textit{// centroids update phase}
	\FOR {$j = 1,2,...,m $}
		\STATE $ (\mu_j^{\phi, (iter)}, \mu_j^{t, (iter)}) \gets $ \textbf{update\_centroid}($j$, $\mathcal{X}$, $\mathcal{M}^{(iter-1)}$, $\mathcal{P}^{(iter-1)}$)
	\ENDFOR

	\STATE $\mathcal{M}^{(iter)} \gets \{\mu_j^{(iter)} | j = 1,2,...,m\}$
	\STATE $\mathcal{P}^{(iter)} \gets \{\mathcal{C}_j^{(iter)} | j = 1,2,...,m\}$ 
\UNTIL {$\mathcal{C}_j^{(iter)} = \mathcal{C}_j^{(iter-1)}, \forall j \in [1,m]$}
\end{algorithmic}
\end{algorithm}

In the centroid update phase, the \textbf{update\_centroid} function recalculates the cluster centroids using the observations in $\mathcal{X}$ and the assignment at the previous iteration.
Therefore the contribution of each cluster to the $\mathcal{J}$ function is minimized.
Each of the temporal and the multidimensional components is calculated individually.
In order to find the values that minimize the objective function, we need to solve the equations:
\begin{equation} \label{eq:derivatives}
	\frac{\partial \mathcal{J}}{\partial \mu_j^d} = 0 \; ; \; \;
	\frac{\partial \mathcal{J}}{\partial \mu_j^t} = 0
\end{equation}
By replacing equations~(\ref{eq:temp-distance}) and~(\ref{eq:penalty-function}) in~(\ref{eq:obj-function}), we obtain the following formula for the objective function:
\begin{align} \label{eq:objective-function-complete}
 	\mathcal{J} = |\mathcal{X}| &- \sum_{j = 1}^{m} \sum_{x_i \in \mathcal{C}_{j}} \left[ \left(1 - \frac{||x_i^d - \mu_j^d||^2}{\Delta x_{max}^2}\right)\left(1 - \frac{||x_i^t - \mu_j^t||^2}{\Delta t_{max}^{2}} \right) \right] \notag \\
	&+ \sum_{x_i \in \mathcal{X}}\sum_{x_k \not\in \mathcal{C}_j} \beta * e^{-\frac{1}{2} \left( \frac{||x_i^t - x_k^t||}{\delta} \right)^2} \mathbbm{1} \left[x_i^\phi = x_k^\phi \right]
\end{align}
Therefore, from equations~(\ref{eq:derivatives}) and~(\ref{eq:objective-function-complete}), we obtain the centroid update formulas:
\begin{equation} \label{eq:centroid-update}
	\mu_j^d = \frac{\sum_{x_i \in \mathcal{C}_j} x_i^d \times \left(1 - \frac{||x_i^t - \mu_j^t||^2}{\Delta t_{max}^{2}} \right)}{\sum_{x_i \in \mathcal{C}_j} \left(1 - \frac{||x_i^t - \mu_j^t||^2}{\Delta t_{max}^{2}} \right)} \; ; \; \;
	\mu_j^t = \frac{\sum_{x_i \in \mathcal{C}_j} x_i^t \times \left(1 - \frac{||x_i^d - \mu_j^d||^2}{\Delta x_{max}^{2}} \right)}{\sum_{x_i \in \mathcal{C}_j} \left(1 - \frac{||x_i^d - \mu_j^d||^2}{\Delta x_{max}^{2}} \right)} 
\end{equation}

Just like the centroid update phase in \KMeans{}, the new centroids in \TDCKMeans{} are also averages over the observations.
Unlike \KMeans{}, the averages are weighted for each component, using the distance from the other.
For example, each observation contributes to the multidimensional description of the new centroid, proportional with its temporal centrality in the cluster.
Observations that are more distant in time (from the centroid) contribute less to the multidimensional description than the ones being closer in time.
A similar logic applies to the temporal component.
The consequence is that the new clusters are coherent both in the multidimensional space and in the temporal one.

\paragraph*{Algorithm's complexity} Equation~(\ref{eq:objective-function-complete}) shows that \TDCKMeans{}' complexity is $\mathcal{O}(n^{2}m)$, due to the penalty term.
Still, the equation can be rewritten, so that only observations belonging to the same entity are tested.
If $p$ is the number of entities and $q$ is the maximum number of observations associated with each entity, then $n = p \times q$.
The complexity of \TDCKMeans{} is $\mathcal{O}(pq^{2}m)$, which is well adapted to Social Science and Humanities datasets, where often a large number of individuals is studied over a relatively short period of time ($p > q$).

\subsection{Fine-tuning the ratio between components}
\label{subsec:measure-tuning-alpha}

The temporal-aware dissimilarity measure, as presented in Equation~(\ref{eq:temp-distance}), gives equal importance to both the multidimensional component and the temporal component.
This might pose problems when the data are not uniformly distributed both in the multidimensional descriptive space and in the temporal space.
If the medium standard deviation reported to the medium distance between pairs of observations is greater in one space than in the other, giving equal weight to the components can lead to important bias in the clustering process.
\textit{E.g.} observations that are very uniformly distributed in the temporal space (same number of observations for each timestamp) and, at the same time, rather compactly distributed in the description space.
In this case, in average, the temporal component weight more in the dissimilarity measure than the multidimensional component.
Consequently, the clustering is biased towards the temporal cohesion of clusters.
Similarly, in some applications, it is desirable to privilege one component over the other.
\textit{E.g.} on a large enough scale, user roles in social networks have a temporal component (new types of roles might appear over the years).
But in a limited time span, it is perfectly acceptable that the roles can coexist simultaneously.
Therefore, the temporal component should have only a mild impact on the overall measure.

We adjust the ratio between the two components by using two tuning factors $\gamma_d$ and $\gamma_t$.
$\gamma_d$ weights the multidimensional component of the temporal-aware dissimilarity measure, whereas $\gamma_t$ weights the temporal component.
Equation~(\ref{eq:temp-distance}) can be rewritten as:
\begin{equation} \label{eq:tuned-temp-distance}
	|| x_i - x_j||_{TA} = 1 - \left(1 - \gamma_d \frac{||x_i^d - x_j^d||^2}{\Delta x_{max}^2}\right)\left(1 - \gamma_t \frac{||x_i^t - x_j^t||^2}{\Delta t_{max}^{2}}\right) 
\end{equation}

When the tuning factor for a certain component is set at zero, the respective component does not contribute to the temporal-aware measure.
When the tuning factor is set to one, no penalty is inflicted to the contribution of the respective component to the measure.
It is immediate that equation~(\ref{eq:temp-distance}) is a special case of equation~(\ref{eq:tuned-temp-distance}), with $\gamma_d = 1$ and $\gamma_t = 1$ (no weights).

\paragraph{Setting the weights $\gamma_d$ and $\gamma_t$}

\begin{figure}[!t]
	\centering
	\includegraphics[width=0.45\textwidth]{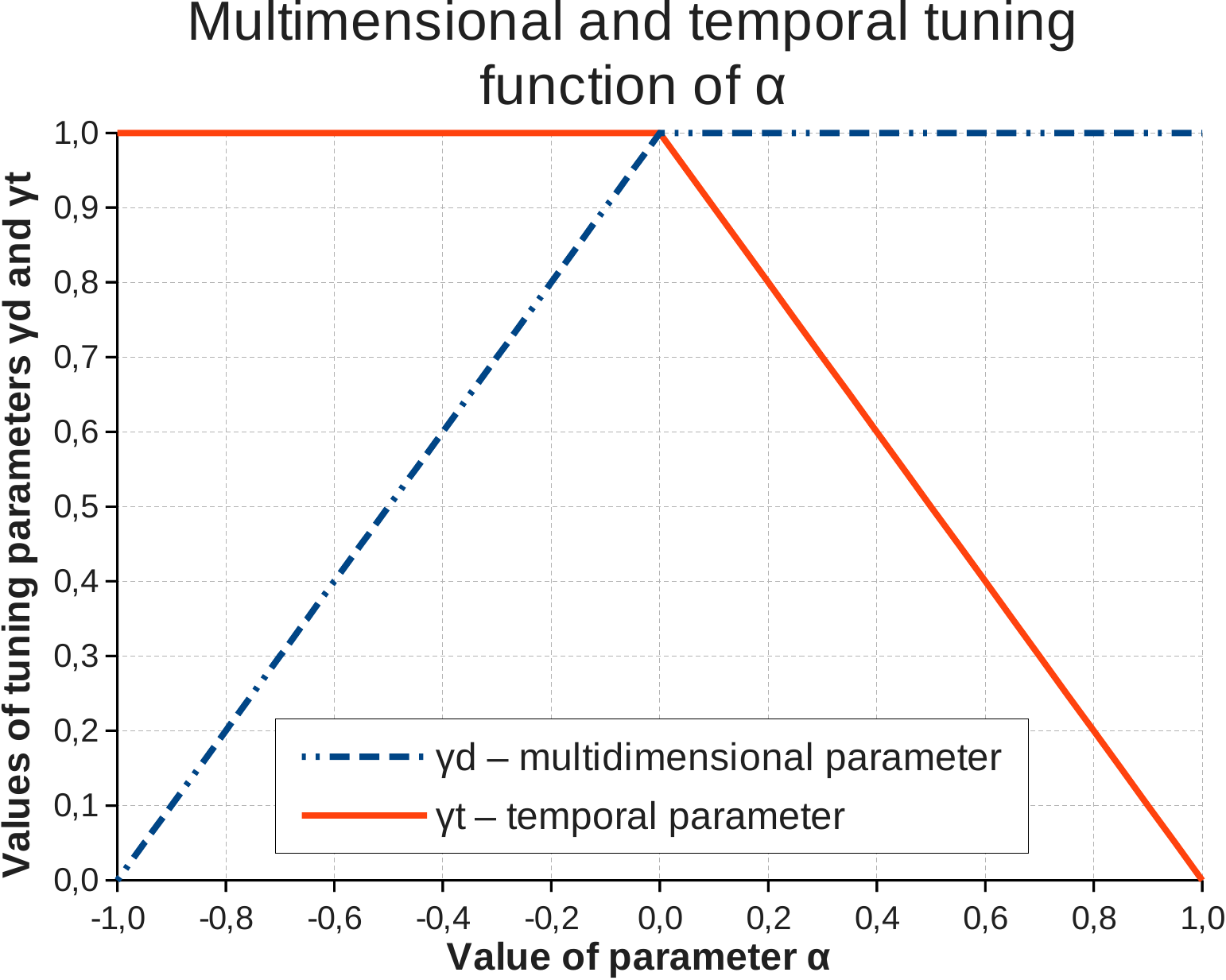}
	\caption{Multidimensional component, temporal component and temporal-aware dissimilarity measure function of $\alpha$}
	\label{fig:alpha-parameter}
	\vspace{-0.1in}
\end{figure}

\begin{figure}[!t]
\centering
	\subfloat[] {
		\includegraphics[width=0.45\textwidth]{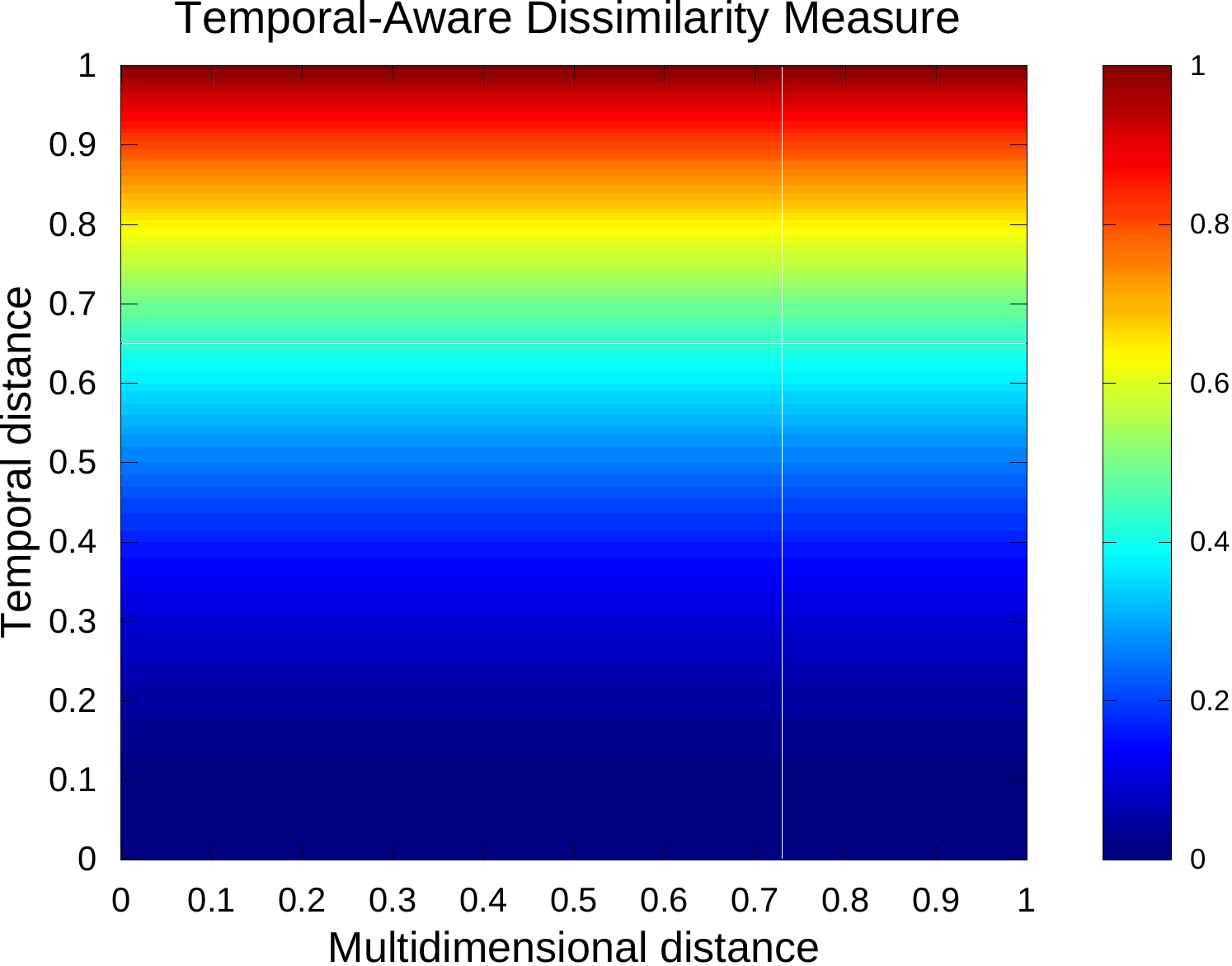}%
		\label{subfig:alpha--1}
	}
	\hfill
	\subfloat[]{
		\includegraphics[width=0.45\textwidth]{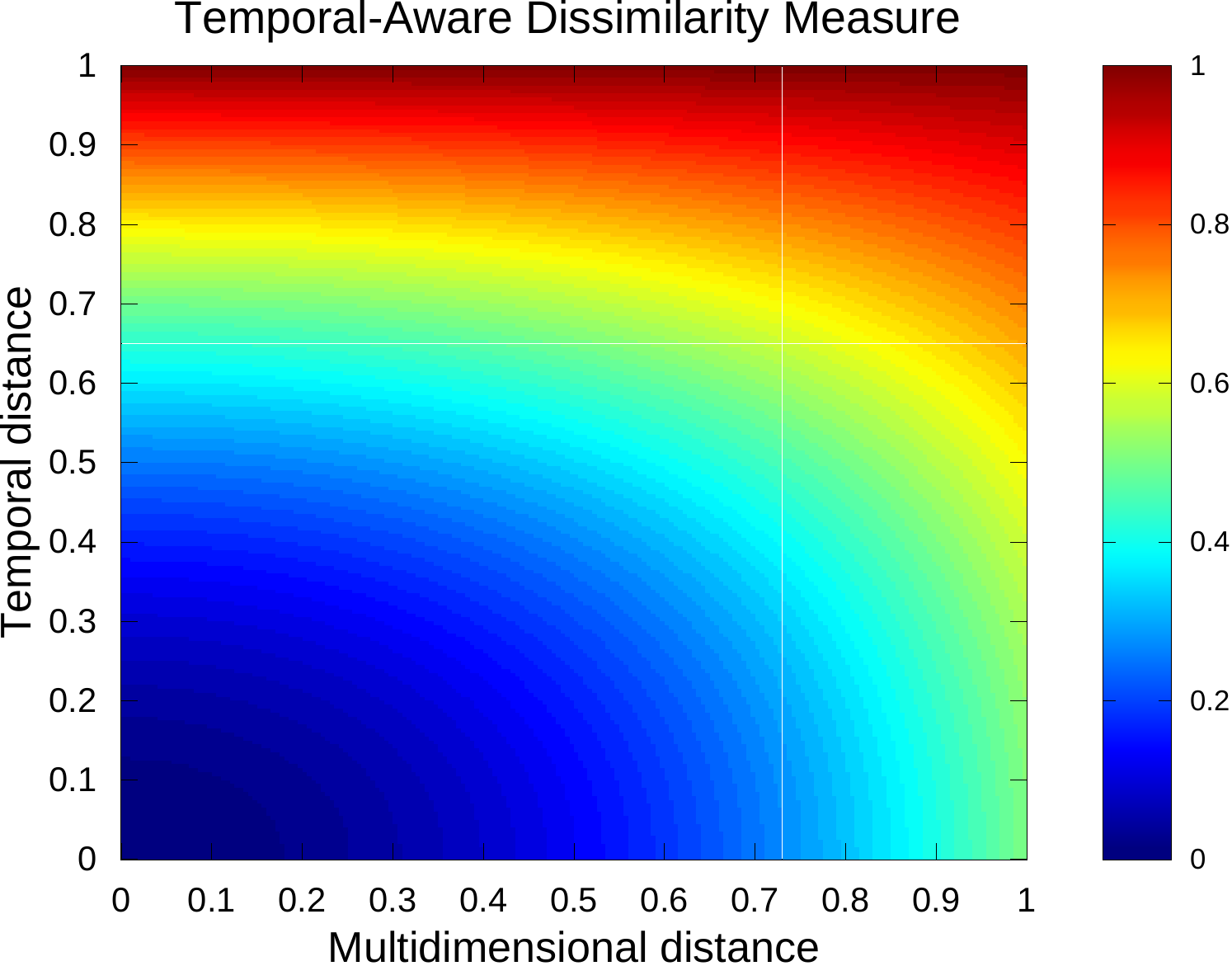}%
		\label{subfig:alpha--0.5}
	}
	\hfill
	\subfloat[]{
		\includegraphics[width=0.45\textwidth]{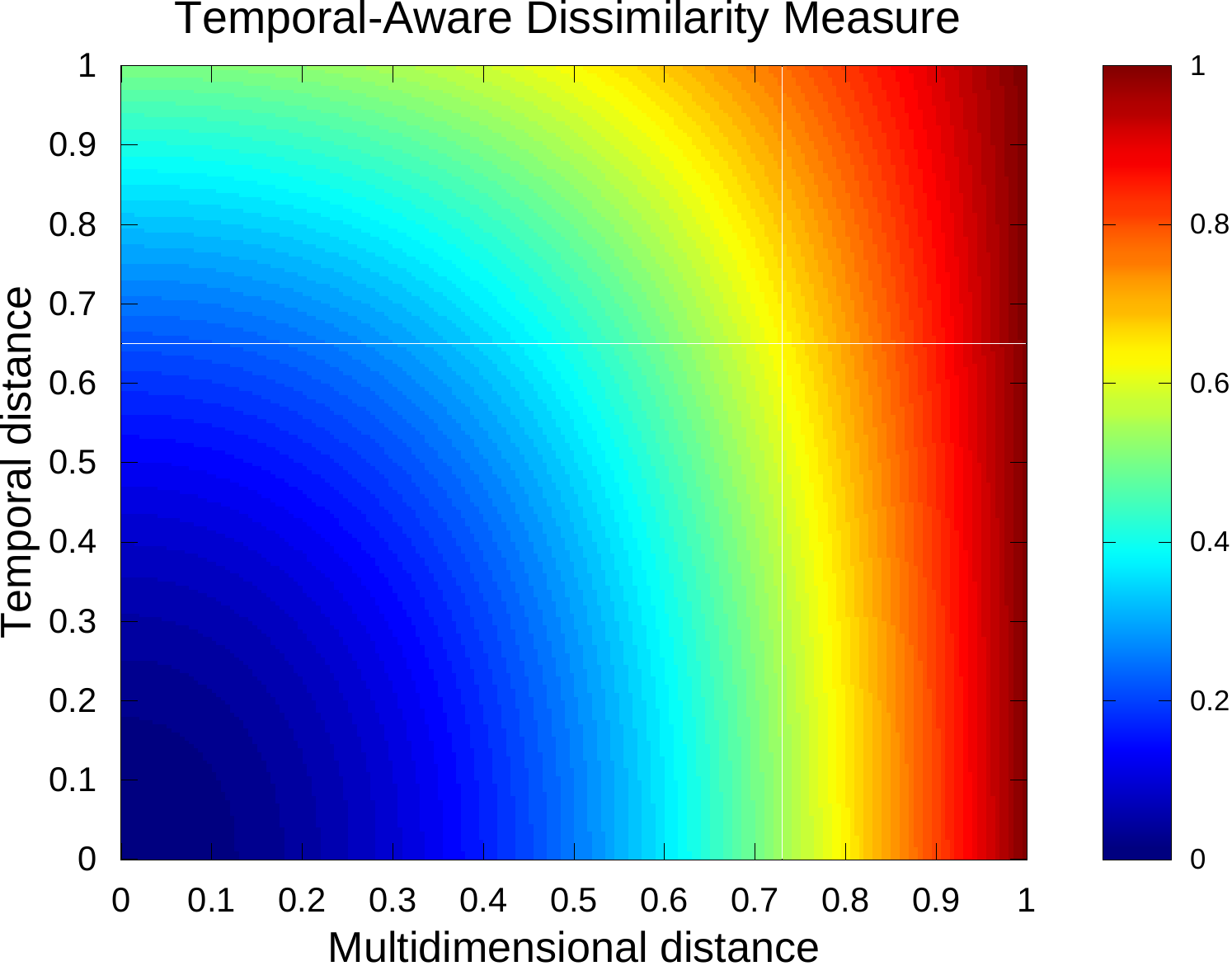}%
		\label{subfig:alpha-0.5}
	}
	\hfill
	\subfloat[]{
		\includegraphics[width=0.45\textwidth]{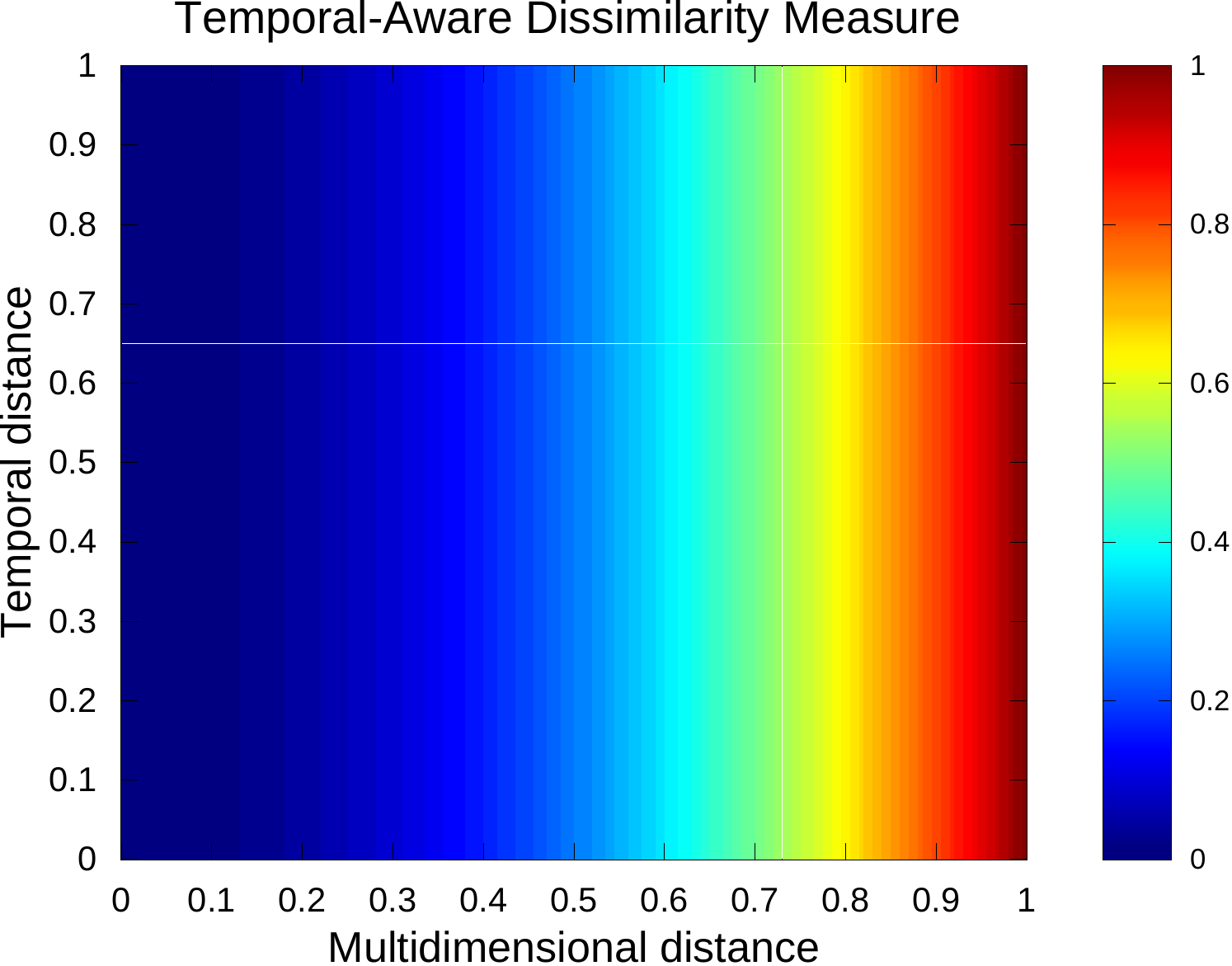}%
		\label{subfig:alpha-1}
	}

	\caption{Color map of the temporal-aware dissimilarity measure for $\alpha=-1$ (a), $\alpha=-0.5$ (b), $\alpha=0.5$ (c) and $\alpha=1$ (d) .}
	\label{fig:colormap-multiple-alpha}
	\vspace{-0.2in}
\end{figure}

$\gamma_d$ and $\gamma_t$ are not independent one from another, their values are set using a unique parameter $\alpha$.
\begin{equation} \label{eq:alpha-fine-tune}
    \gamma_d= 
\begin{cases}
    1 + \alpha ,& \text{if } \alpha \leq 0 \\
    1, 			& \text{if } \alpha > 0
\end{cases} 
	\;\;;\;\;
	\gamma_t= 
\begin{cases}
    1 ,		   & \text{if } \alpha \leq 0 \\
    1 - \alpha,& \text{if } \alpha > 0
\end{cases} 
\end{equation}
$\alpha$ acts as a slider, taking values from $-1$ to $1$.
Figure~\ref{fig:alpha-parameter} shows the evolution $\gamma_d$ and $\gamma_t$ with $\alpha$.
Also, Figure~\ref{fig:colormap-multiple-alpha} shows the color map of the temporal-aware dissimilarity measure for multiple values of $\alpha$.

When $\alpha=-1$, then $\gamma_d=0$ and $\gamma_t=1$.
The multidimensional component is eliminated and only the time difference between the two observations is considered.
The temporal-aware measure becomes a normalized time difference ($ || x_i - x_j ||_{TA} = \frac{||x_i^t - x_j^t||^2}{\Delta t_{max}^{2}} $).
The color map in Figure~\ref{subfig:alpha--1} ($\alpha=-1$) shows that the values of the dissimilarity measure is independent of the multidimensional component.

As the value of $\alpha$ increases, the weight of the descriptive component increases as well.
In Figure~\ref{subfig:alpha--0.5} ($\alpha=-0.5$), the multidimensional component has a limited impact on the overall measure.
When $\alpha=0$, then $\gamma_d=1$ and $\gamma_t=1$, both components have equal importance, as proposed initially in Equation~(\ref{eq:temp-distance}).
In Figure~\ref{subfig:alpha-0.5} ($\alpha=0.5$), the color map shows that the multidimensional component has a larger impact then the temporal component.
Large values of the temporal component have only moderate influence over the measure.
When $\alpha=1$ (color map in Figure~\ref{subfig:alpha-1}), then $\gamma_d=1$ and $\gamma_t=0$, the temporal dimension is eliminated and the measure becomes a normalized Euclidean distance ($ || x_i - x_j ||_{TA} = \frac{||x_i^d - x_j^d||^2}{\Delta x_{max}^{2}} $).

Since the temporal-aware dissimilarity measure is used in the objective function in Equation~(\ref{eq:objective-function-complete}), the later changes accordingly to integrate the tuning factors.
$\gamma_d$ and $\gamma_t$ behave as constants in the derivation formulas in Equation~(\ref{eq:derivatives}).
As a result, the centroid update formulas in Equation~(\ref{eq:centroid-update}) are rewritten as:
\begin{equation*}
	\mu_j^d = \frac{\sum_{x_i \in \mathcal{C}_j} x_i^d \times \left(1 - \gamma_t \frac{||x_i^t - \mu_j^t||^2}{\Delta t_{max}^{2}} \right)}{\sum_{x_i \in \mathcal{C}_j} \left(1 - \gamma_t \frac{||x_i^t - \mu_j^t||^2}{\Delta t_{max}^{2}} \right)} \;\; ; \; \;\mu_j^t = \frac{\sum_{x_i \in \mathcal{C}_j} x_i^t \times \left(1 - \gamma_d \frac{||x_i^d - \mu_j^d||^2}{\Delta x_{max}^{2}} \right)}{\sum_{x_i \in \mathcal{C}_j} \left(1 - \gamma_d \frac{||x_i^d - \mu_j^d||^2}{\Delta x_{max}^{2}} \right)}
\end{equation*}
The tuning between the multidimensional and temporal component in the temporal-aware dissimilarity measure propagates into the centroid update formula of \TDCKMeans{}.
We study, in Section~\ref{subsec:parameters-alpha}, the influence of the tuning parameter and we propose an heuristic to set its value.

\subsection{Inferring a graph structure for the temporal clusters}
\label{subsec:graph-structure}

In Figure~\ref{subfig:cluster-structuring}, when discussing the desired output of our system, we 
presented the obtained temporal clusters under the form of a graph.
The nodes represent the evolution phases and an edge between two nodes $\mu_p$ and $\mu_q$ indicates that the transition $\mu_p \longrightarrow \mu_q$ is part of a typical evolution.
Since each temporal cluster is interpreted as an evolution phase, the visualization under the form of a graph allows quick understanding of how the different phases are organized both (i) in time (phases on the left side of Figure~\ref{subfig:cluster-structuring} have a lower timestamp than those on the right side) and (ii) considering the transitions of the entities through phases.
Intuitively, the strength of the connection between two phases is proportional with the number of entities which present transitions between the two given phases.

We consider that an entity $\phi_l$ presents a transition between $\mu_p$ and $\mu_q$ ($\mu_p \xrightarrow{\phi_l} \mu_q$) if and only if two consecutive observations exist, associated with the given entity, where the first observation (ordered by their timestamp) is clustered under $\mu_p$ and the second one is clustered under $\mu_q$.
Formally:
\begin{align*} 
	x_a, x_b \in \mathcal{D} \text{ consecutive } \Leftrightarrow \text{ } & x_a^{\phi} = x_b^{\phi} = \phi_l, x_a^{t} \leq x_b^{t} \text{ and } \\
	& \nexists x_c \in \mathcal{D}, x_c^{\phi} = \phi_l \text{ so that } x_a^{t} \leq x_c^t \leq x_b^{t}
\end{align*}
\vspace{-0.2in}
\begin{align}	
	\mu_p \xrightarrow{\phi_l} \mu_q \Leftrightarrow
	\exists x_a, x_b \in \mathcal{D} \text{ so that }
	\begin{cases}
    	x_a^{\phi} = x_b^{\phi} = \phi_l \text{ and}\\
    	x_a, x_b \text{ consecutive and } \\
    	x_a \in \mathcal{C}_p , x_b \in \mathcal{C}_q
	\end{cases} \label{eq:transition-definition}
\end{align}
Furthermore, we define the intersection similarity measure between two phases, which is based on the normalized number of entities that present a transition between the two phases.
Formally, we define the $inter_{\phi} (\mu_p, \mu_q)$:
\vspace{-0.1in}
\begin{equation} \label{eq:intersection-measure}
	inter_{\phi} (\mu_p, \mu_q) = \frac{|\{ \phi_l \in \Phi | \mu_p \xrightarrow{\phi_l} \mu_q \}|}{|\Phi|}
\end{equation}
where $inter_{\phi} (\mu_p, \mu_q) \in [0,1]$ and needs to be maximized.

We infer a graph structure between the temporal clusters, by constructing an adjacency matrix using the intersection similarity measure.
The graph construction is performed \textit{a posteriori}, after the temporal clusters are calculated.
We define the adjacency matrix $ A = (a_{p,q})$, where $ a_{p,q} = inter_{\phi} (\mu_p, \mu_q)$.
By replacing equations~\ref{eq:transition-definition} and~\ref{eq:intersection-measure} into this definition, we obtain:
\begin{equation} \label{eq:adjacency-matrix}
	a_{p,q} = \frac{|\{ 1 \leq l \leq |\Phi| | \exists x_a, x_b \in (\mathcal{C}_p,\mathcal{C}_q)  \text{ so that } x_a^{\phi} = x_b^{\phi} = \phi_l \text{ and } x_a^{t}, x_b^{t} \text{ consecutive}\}|}{|\Phi|}
\end{equation}
We construct $A^{*}$, a binary adjacency matrix, by using a threshold $\gamma$:
\vspace{-0.1in}
\begin{equation*}
	A^{*} = (a^{*}_{p,q}) \text{ with } 
	a^{*}_{p,q} = 
	\begin{cases}
    	0 &, \text{ if } a_{p,q} < \gamma \\
    	1 &, \text{ if } a_{p,q} \geq \gamma
	\end{cases}  
\end{equation*}
The filtering parameter $\gamma$ is dataset dependent and automatically setting its value is part of the perspectives of our work.

\section{Experiments}
\label{sec:xp}

\subsection{Dataset}

Experimentations with Time-Driven Constrained \KMeans{} are performed on a dataset issued from political sciences: \textit{Comparative Political Data Set I}~\cite{ARM11}.
It is a collection of political and institutional data, which consists of annual data for 23 democratic countries for the period from 1960 to 2009.
The dataset contains 207 political, demographic, social and economic variables.

The dataset was cleaned by removing redundant variables (\textit{e.g.} country identifier and postal code) and the corpus was preprocessed by removing entity bias from the data.
For example, it is difficult to compare, on the raw data, the evolution of population between populous country and one with fewer inhabitants, since any evolution in the 50 years timespan of the dataset will be rendered meaningless by the initial difference.
Inspired from panel data econometrics~\cite{DOR89}, we remove the entity-specific, time-invariant effects, since we assume them to be fixed over time.
We subtract from each value the average over each attribute and over each entity.
We retain the time-variant component, which is in turn normalized, in order to avoid giving too much importance to certain variables.
The obtained dataset is under the form of triples $(country, year, description)$.

\subsection{Qualitative evaluation}

\begin{figure}[!t]
\centering
	\subfloat[] {
		\includegraphics[height=0.169\textheight]{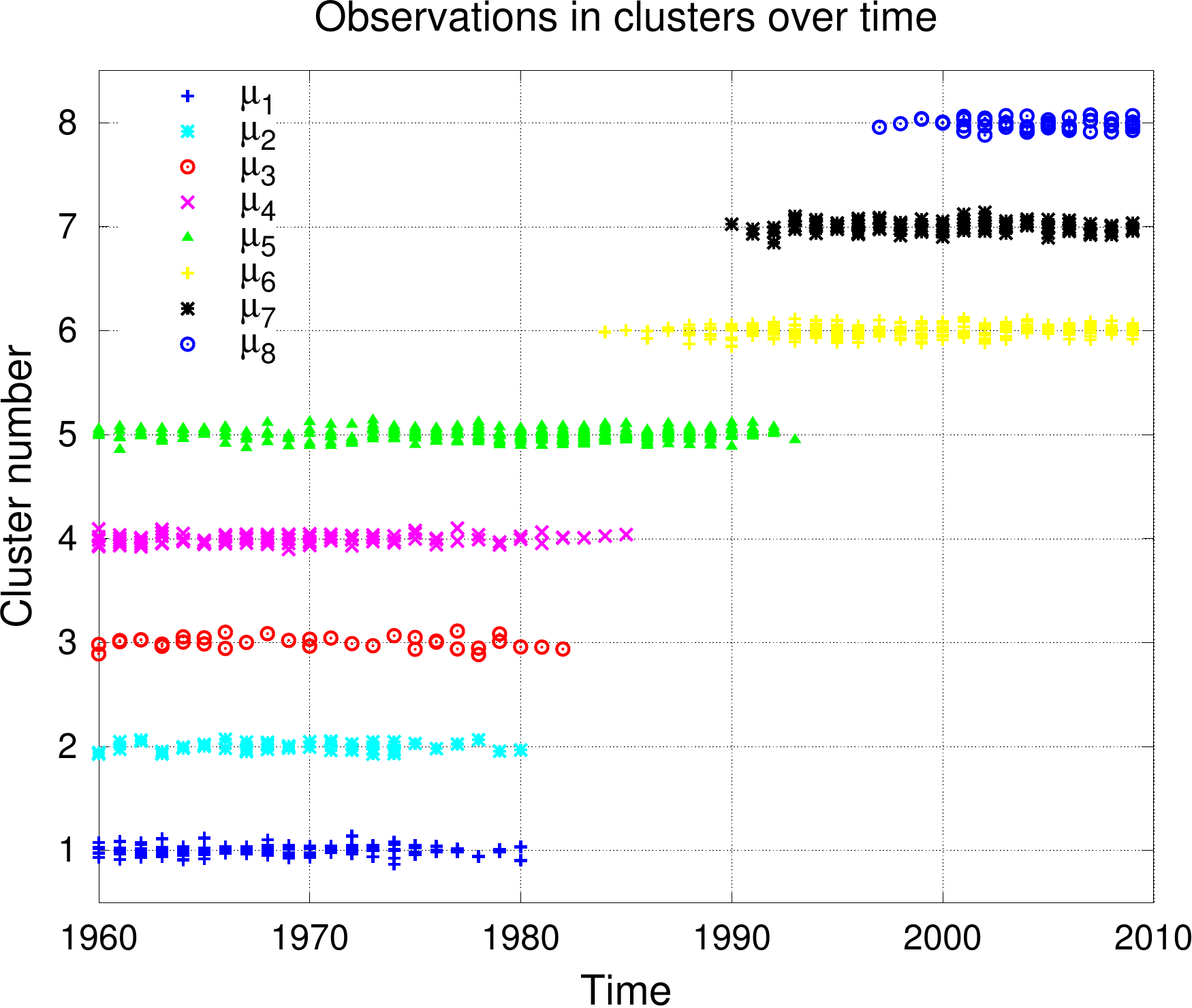}%
		\label{subfig:example-obs-clus-time}
	}
	\hfill
	\subfloat[]{
		\includegraphics[height=0.169\textheight]{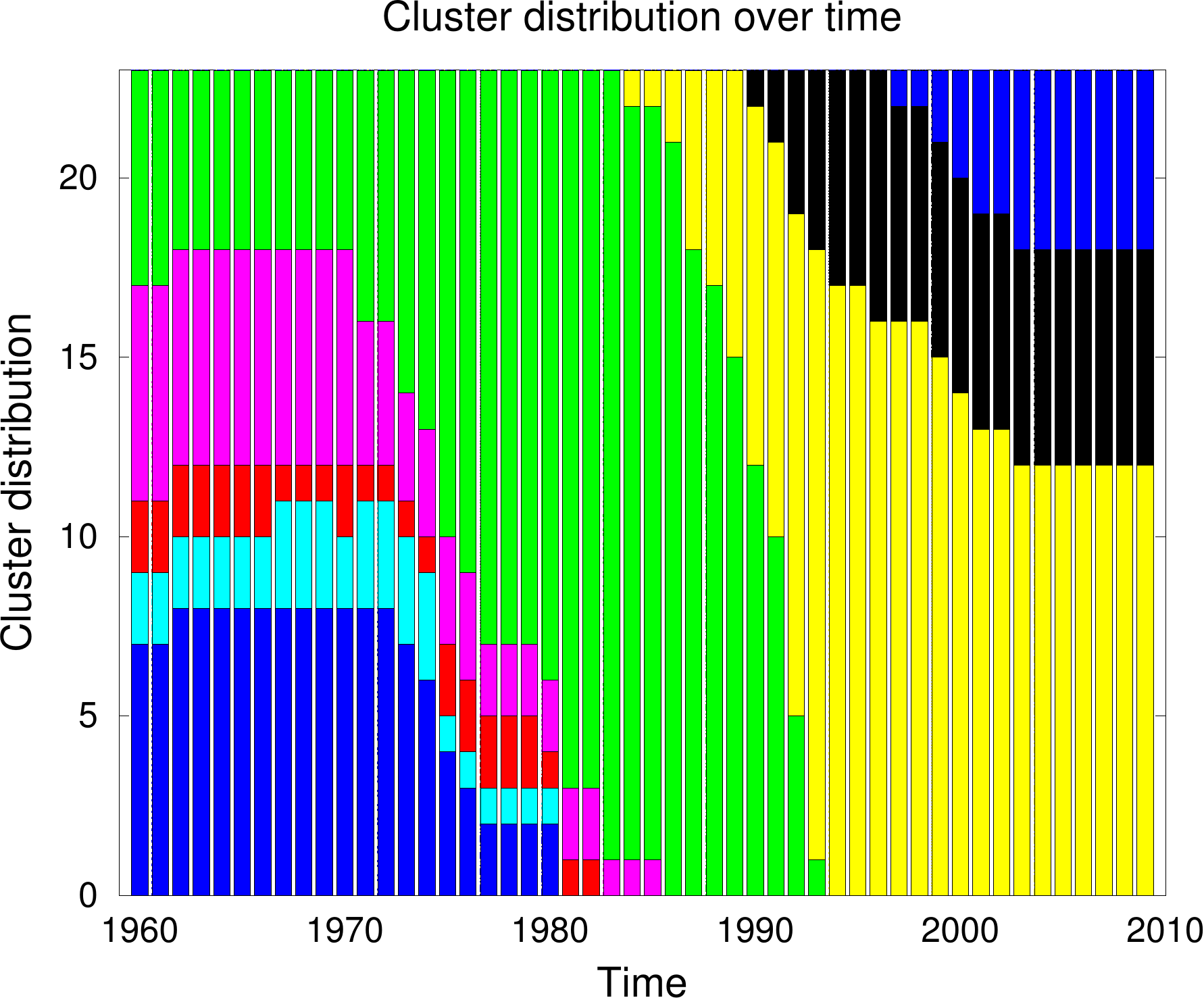}%
		\label{subfig:example-clus-time}
	}
	\hfill
	\subfloat[]{
		\includegraphics[height=0.169\textheight]{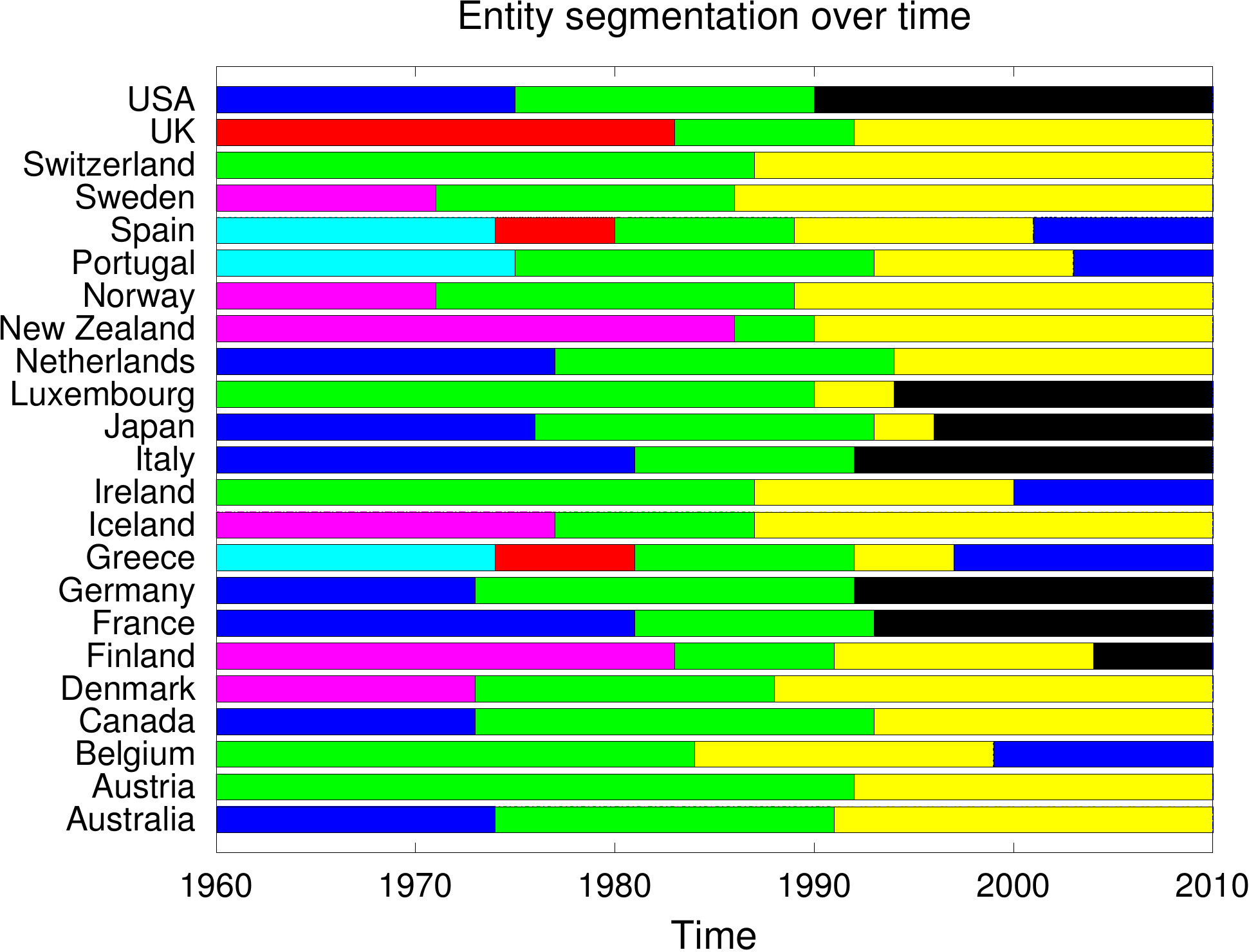}%
		\label{subfig:example-entity-seg}
	}

	\caption{Typical evolution patterns constructed by \TDCKMeans{} on \textit{Comparative Political Data Set I} with 8 clusters. The distribution over time of observations in each cluster (a), how many entities belong in a certain clusters for each year (b) and the segmentation of entities over clusters (c).}
	\label{fig:example-run}
	\vspace{-0.2in}
\end{figure}

When studying the evolution of countries over the years, it is quite obvious for the human reader why the evolutions of the eastern European countries resemble each other for most of the second half of the twentieth century.
The reader would create a group entitled ``Communism'', extending from right after the Second World War until roughly 1990, for defining the typical evolution of communist countries.
One would expect that, based on a political dataset, the algorithms would succeed in identifying such typical evolutions and segment the time series of each of these countries accordingly.
Figure~\ref{fig:example-run} shows the typical evolution patterns constructed by \TDCKMeans{} (with $\beta = 0.003$ and $\delta = 3$, obtained as shows in Section~\ref{subsec:parameters-beta-delta}), when asked for 8 clusters.
The distribution over time of observations in each cluster is given in Figure~\ref{subfig:example-obs-clus-time}.
All constructed clusters are fairly compact in time and have limited temporal extents.
They can be divided into two temporal groups.
In the first one, clusters $\mu_1$ to $\mu_5$ consistently overlap.
Same for clusters $\mu_6$ to $\mu_8$, in the second group.
This indicates that the evolution of each country passes by at least one cluster from each group.
The turning point between the two groups is around 1990.
Figure~\ref{subfig:example-clus-time} shows how many countries belong in a certain cluster for each year.
Clusters $\mu_5$ and $\mu_6$ contain most of the observations, suggesting the general typical evolution.

The meaning of each constructed cluster starts to unravel only when studying the segmentation of countries over clusters, in Figure~\ref{subfig:example-entity-seg}.
For example, cluster $\mu_2$ regroups the observations belonging to Spain, Portugal and Greece from 1960 up until around 1975.
Historically, this coincides with the non-democratic regimes in those countries (Franco's dictatorship in Spain, the ``Regime of the Colonels'' in Greece).
Likewise, cluster $\mu_4$ contains observations of countries like Denmark, Finland, Iceland, Norway, Sweden and New Zealand.
This cluster can be interpreted as the ``Swedish Social and Economical Model'' of the Nordic countries, to which the algorithm added, interestingly enough, New Zealand.
In the second period, cluster $\mu_8$ regroups observations of Greece, Ireland, Spain, Portugal and Belgium, the countries which seemed the most fragile in the aftermaths of the economical crises of 2008.

\begin{figure}[htb]
\centering
	\subfloat[] {
		\includegraphics[width=0.99\textwidth]{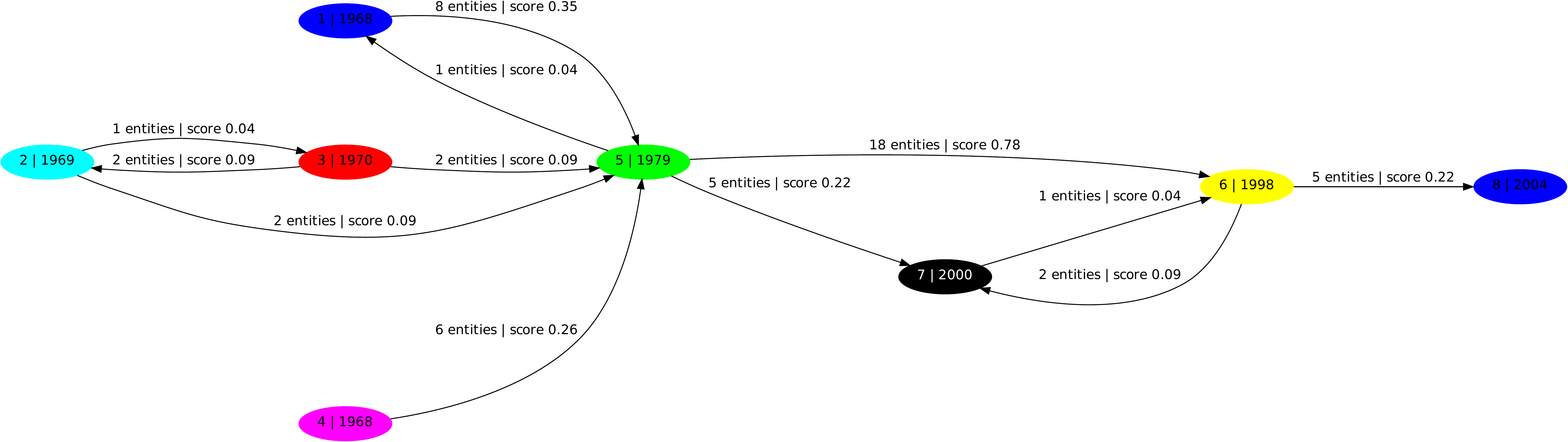}%
		\label{subfig:graph-complete}
	}
	\hfill
	\subfloat[]{
		\includegraphics[width=0.9\textwidth]{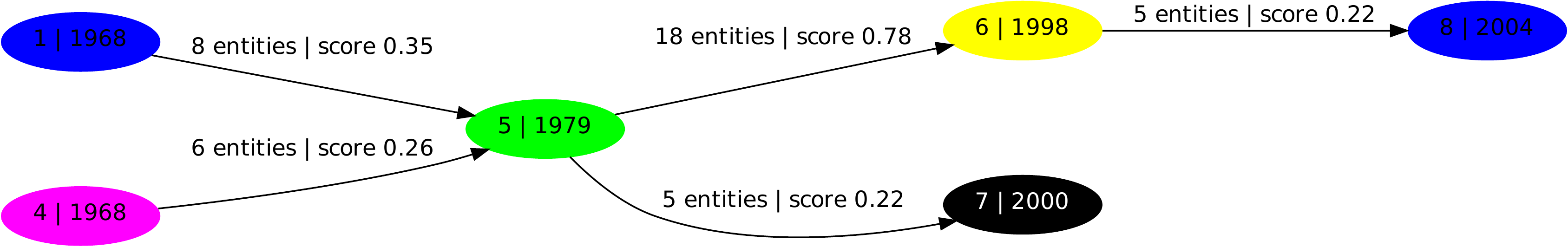}%
		\label{subfig:graph-filtered}
	}

	\caption{Structuring the temporal clusters as a graph, without filtering ($\gamma = 0$) (a) and filtered with $\gamma = 0.2$ (b).}
	\label{fig:graph-structure}
	\vspace{-0.1in}
\end{figure}

Similar conclusions can be drawn from the constructed graph structure, presented in Figure~\ref{fig:graph-structure}.
Each temporal cluster is represented as a node and the scores indicated on each edge are calculated as shown in Equation~\ref{eq:adjacency-matrix}.
The graph containing all transitions are represented in Figure~\ref{subfig:graph-complete} (no filtering, $\gamma = 0$).
We obtain a graph containing more general evolutions, by filtering with the threshold $\gamma = 0.2$.
Some ``rare'' phases completely disappear (\textit{i.e.}, phases $\mu_2$ and $\mu_3$) together with some of the arcs.
We recognize in the resulted graph, shown in Figure~\ref{subfig:graph-filtered}, some of the evolutions identified earlier.
The evolution $\mu_4 \longrightarrow \mu_5 \longrightarrow \mu_6$ corresponds to the ``Swedish Social and Economical Model'', whereas the evolution $\mu_5 \longrightarrow \mu_6 \longrightarrow \mu_8$ identifies the fragile European economies of the 2008 economical crises.
From the filtered evolution graph, another typical evolution emerges: $\mu_1 \longrightarrow \mu_5 \longrightarrow \mu_7$, which is present for countries as USA, Germany, Italy and France.
We interpret this evolution as that of countries with stable social and economical environments.

\subsection{Evaluation measures}
\label{subsec:xp-evaluation-measures}

Since the dataset contains no labels to report to as ground truth, we use the classical Information Theory measures in order to numerically evaluate the proposed algorithms.
We evaluate separately each of the three goals that we propose in Section~\ref{sec:introduction}.

\textbf{Create clusters that are coherent in the multidimensional description space.}
It is desirable that observations that have similar multidimensional descriptions to be partitioned under the same cluster.
The similarity in the description space is measured by the multidimensional component of the temporal-aware dissimilarity measure.
This goal is pursued by all classical clustering algorithms (like \KMeans{}) and any traditional clustering evaluation measure~\cite{HAL01} can be used to asses it.
We choose the mean cluster variance, which is traditionally used in clustering to quantify the dispersion of observations in clusters.
The \textit{MDvar} measure is defined as:
\vspace*{-0.22cm}
\begin{equation*}
	MDvar = \frac{1}{|\mathcal{X}|} \times \sum_{j = 1}^{m} \sum_{x_i \in \mathcal{C}_{j}} ||x_i^d - \mu_j^d||^2
\end{equation*}

\textbf{Create temporally coherent clusters, with limited extend in time.}
This goal is very similar to the previous one, translated in the temporal space.
It is desirable that observations that are assigned to the same cluster to be similar in the temporal space (\textit{i.e.} to be close in time).
The similarity in the temporal space is measured by the temporal component in the temporal-aware dissimilarity measure.
The limited time extent of a centroid implies small temporal distances between observations timestamp and the centroid timestamp.
As a result, the variance can also be used to measure the dispersion of clusters in the temporal space.
Similarly to \textit{MDvar}, the \textit{Tvar} measure is defined as:
\vspace*{-0.22cm}
\begin{equation*}
	Tvar = \frac{1}{|\mathcal{X}|} \times \sum_{j = 1}^{m} \sum_{x_i \in \mathcal{C}_{j}} ||x_i^t - \mu_j^t||^2
\end{equation*}

\begin{figure}[ht]
	\centering
	\includegraphics[width=0.7\textwidth]{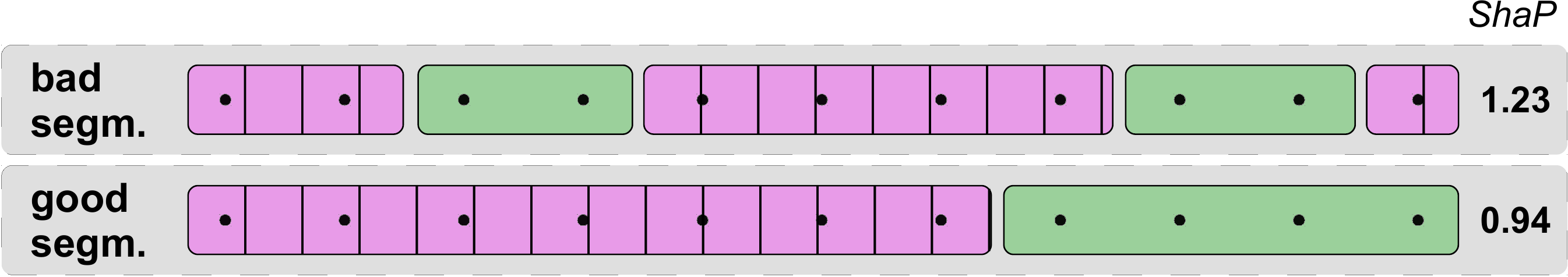}
	\caption{Examples of a good and a bad segmentation in contiguous chunks and their related \textit{ShaP} score.}
	\label{fig:good-bad-shap}
	\vspace{-0.18in}
\end{figure}

\textbf{Segment the temporal series of observations of each entity into a relatively small number of contiguous segments.}
The goal is to have successive observations belonging to an entity grouped together, rather that scattered in different clusters.
The Shannon entropy can quantify the number of clusters which regroup the observations of an entity, but it is insensible to alternations between two classes (evolutions like $\mu_1 \longrightarrow \mu_2 \longrightarrow \mu_1 \longrightarrow \mu_2 $).
We evaluate using an adapted mean Shannon entropy of clusters over entities, which weights the entropy by a penalty factor depending on the number of continuous segments in the series of each entity.
The \textit{ShaP} measure is calculated as:
\vspace*{-0.22cm}
\begin{equation*}
	ShaP = \frac{1}{|\mathcal{X}|} \times \sum_{x_i \in \mathcal{X}} \sum_{j=1}^m \left( -p(\mu_j)\times \log_2(p(\mu_j)) \times \left( 1 + \frac{n_{ch} - n_{min}}{n_{obs} - 1}\right) \right)
\end{equation*}
where $n_{ch}$ is the number of changes in the cluster assignment series of an entity, $n_{min}$ is the minimal required number of changes and $n_{obs}$ is the number of observations for an entity.
For example, in Figure~\ref{fig:good-bad-shap}, if the series of 11 observations of an entity is assigned to two clusters, but it presents 4 changes, the entropy penalty factor will be $1 + \frac{4 - 1}{11 - 1} = 1.33$.
The \textit{ShaP} score for this segmentation will be $1.23$, compared to a score of $0.94$ of the ``ideal'' segmentation (only two contiguous chunks).

The ``ideal'' values for \textit{MDvar}, \textit{Tvar} and \textit{ShaP} is zero and, in all of the experiments presented in the following sections, we search to minimize the values of the three measures.

\subsection{Quantitative evaluation}
\label{subsec:quantitative-evaluation}

\begin{figure}[!t]
	\centering
	\subfloat[] {
		\includegraphics[width=0.31\textwidth]{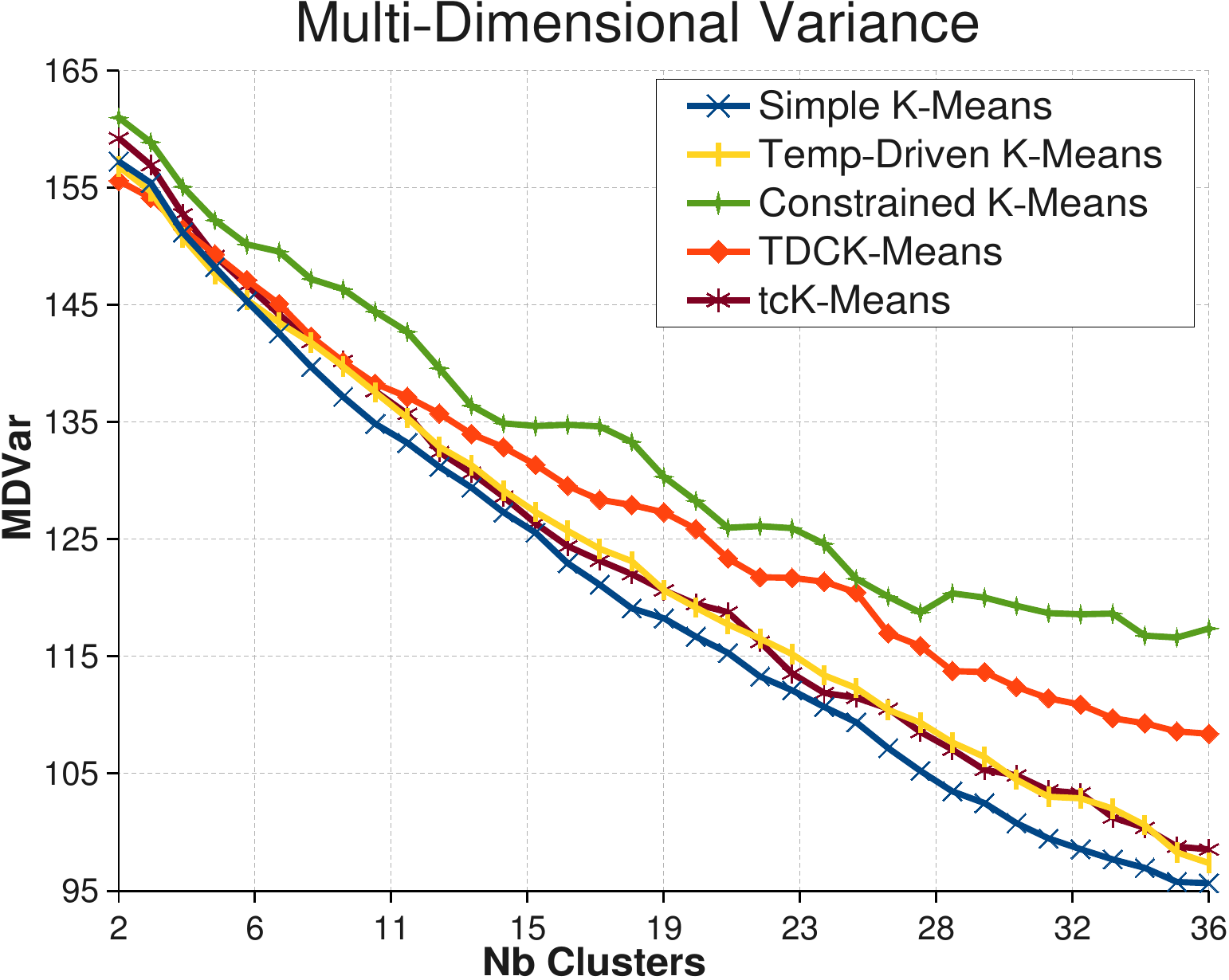}%
		\label{subfig:mdvar-noclus}
	}
	\hfill
	\subfloat[]{
		\includegraphics[width=0.31\textwidth]{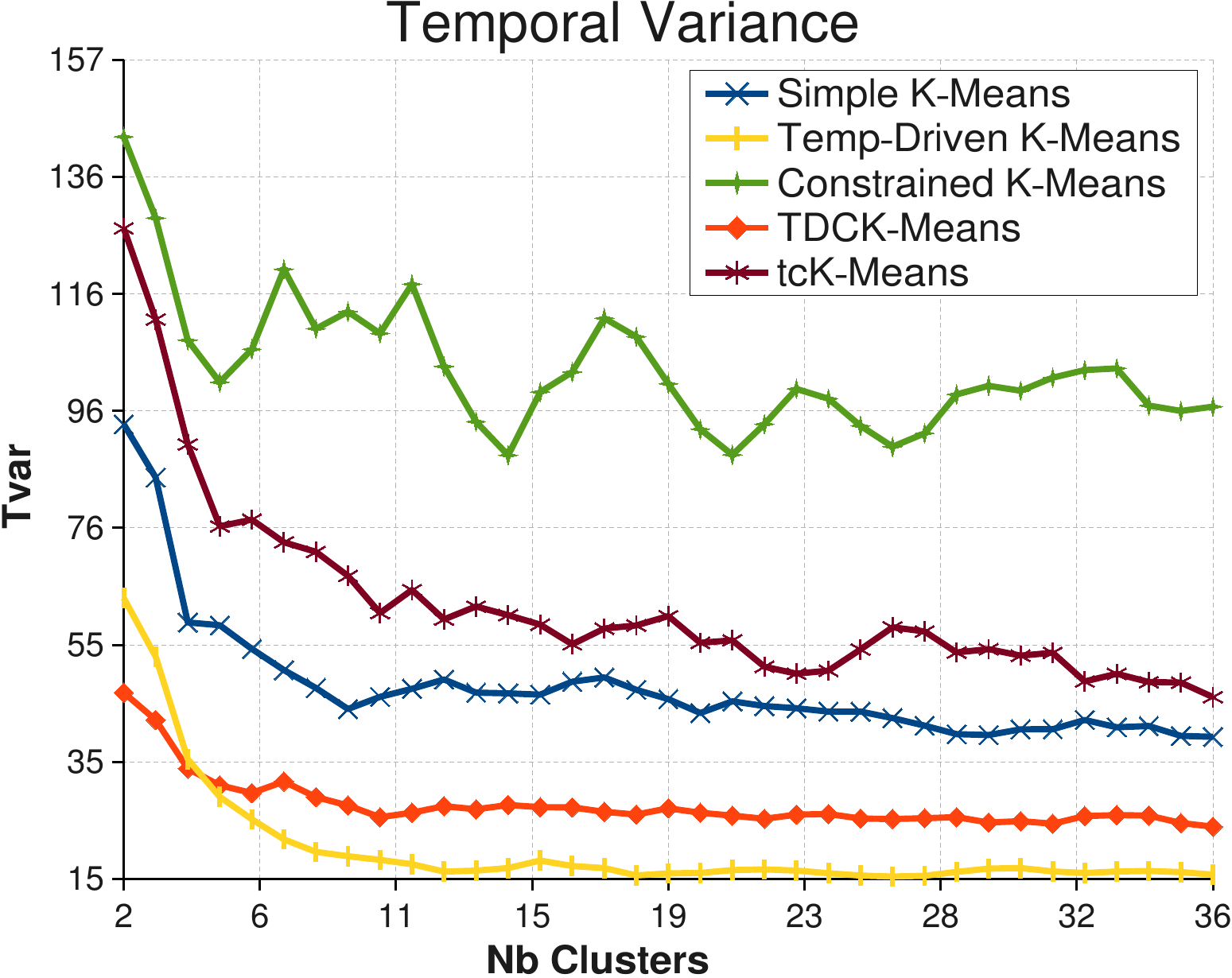}%
		\label{subfig:tvar-noclus}
	}
	\hfill
	\subfloat[]{
		\includegraphics[width=0.31\textwidth]{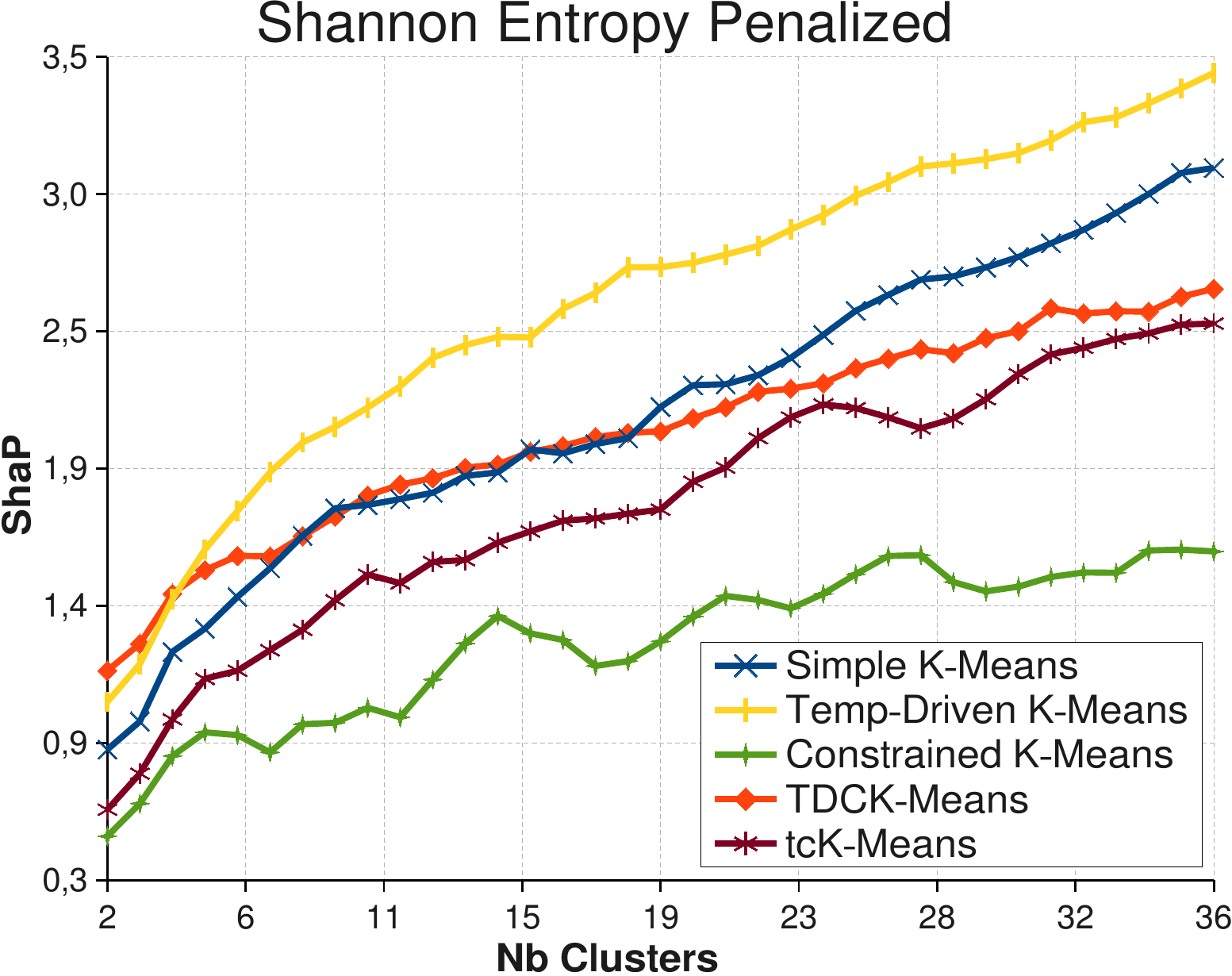}%
		\label{subfig:shap-noclus}
	}

	\caption{\textit{MDvar} (a), \textit{Tvar} (b) and \textit{ShaP} (c) values of the considered algorithms when varying the number of clusters.}
	\label{fig:measures-clus}
	\vspace{-0.2in}
\end{figure}

For each combination of algorithms and parameters, we execute 10 times and compute only the average and the standard deviation.
We vary $m$, the number of clusters, from 2 to 36.
The performances of five algorithms are compared from a quantitative point of view:
\squishlisttwo
	\item \textbf{Simple \KMeans{}} - clusters the observations based solely on their resemblance in the multidimensional space;
	
	\item \textbf{Temporal-Driven \KMeans{}} - optimizes only the temporal and multidimensional components, without any contiguity constraints;
	combines \KMeans{} with the temporal-aware dissimilarity measure defined in Section~\ref{subsec:proposal-temporal-dissimilarity-measure}.
	Parameters: $\alpha=0$ ($\alpha$ defined in Equation~\ref{eq:centroid-update}) and $\beta=0$ ($\beta$ defined in Equation~\ref{eq:penalty-function});
	
	\item \textbf{Constrained \KMeans{}} - uses only the multidimensional space (and not the temporal component) together with the penalty component, as proposed in Section~\ref{subsec:proposal-penalty-function}. 
	Parameters: $\alpha=1$, $\beta = 0.003$ and $\delta = 3$ ($\delta$ defined in Equation~\ref{eq:penalty-function});
	
	\item \textbf{\TDCKMeans{}} - the Temporal-Driven Constrained Clustering algorithm proposed in Section~\ref{subsec:proposal-tdck-means}. 
	$\alpha=0$, $\beta = 0.003$ and $\delta = 3$;
	
	\item \textbf{tcK-Means} - the temporal constrained clustering algorithm proposed in~\cite{LIN06}. 
	It uses a threshold penalty function $w(x_i^{t_i},x_j^{t_i}) = \alpha^* \mathbbm{1} (|x_i^t - x_j^t| < d^*)$ when observations $ x_i$ and $x_j$ are not assigned to the same cluster. 
	We adapt it to the multi-entity case by applying it only to observations belonging to the same entity. 
	Parameters: $\alpha^* = 2, d^* = 4$.
\squishend
The $\alpha^*$ parameter in \textbf{tcK-Means} should not be mistaken with the $\alpha$ parameter in \textbf{\TDCKMeans{}}, as they do not have the same meaning.
	In \textbf{tcK-Means}, $\alpha^*$ controls the weight of the penalty function, whereas in \textbf{\TDCKMeans{}} $\alpha$ is the fine-tuning parameter.

\begin{table}[t]
\caption{Mean values for indicators and standard deviations}
\label{tab:result-all}

\small
\centering
\begin{tabular}{llrr|rr|rr}
\toprule
 & \multicolumn{1}{c}{\textit{Algorithm}} &  \multicolumn{2}{c}{\textit{MDvar}} & \multicolumn{2}{c}{\textit{Tvar}} & \multicolumn{2}{c}{\textit{ShaP}} \\ \midrule
 
\multirow{4}{*}{\begin{sideways}\texttt{Scores}\end{sideways}} 
& \textbf{Simple \KMeans{}} & \textbf{120.59} & \textit{2.97} & 48.01 & \textit{8.87} & 2.15 & \textit{0.23} \\ 
& \textbf{Temp-Driven \KMeans{}} & 122.98 & \textit{2.85} & \textbf{19.97} & \textit{5.39} & 2.58 & \textit{0.18} \\ 
& \textbf{Constrained \KMeans{}} & 132.69 & \textit{8.07} & 103.15 & \textit{42.98} & \textbf{1.24} & \textit{0.5} \\ 
& \textbf{\TDCKMeans{}} & 127.81 & \textit{3.96} & 27.54 & \textit{5.81} & 2.06 & \textit{0.2} \\ 
& \textbf{tcK-Means} & 123,04 & \textit{3.8} & 62.44 & \textit{24.16} & 1.79 & \textit{0.32} \\ 

\midrule

\multirow{3}{*}{\begin{sideways}\texttt{\% Gain}\end{sideways}} 
& \textbf{Temp-Driven \KMeans{}} & \multicolumn{2}{c|}{-1.99\%} & \multicolumn{2}{c|}{\textbf{58.40\%}} & \multicolumn{2}{c}{-19.63\%} \\ 
& \textbf{Constrained \KMeans{}} & \multicolumn{2}{c|}{-10.04\%} & \multicolumn{2}{c|}{-114.84\%} & \multicolumn{2}{c}{\textbf{42.21\%}} \\ 
& \textbf{\TDCKMeans{}} & \multicolumn{2}{c|}{-5.99\%} & \multicolumn{2}{c|}{42.64\%} & \multicolumn{2}{c}{4.19\%} \\ 
& \textbf{tcK-Means} & \multicolumn{2}{c|}{-2.03\%} & \multicolumn{2}{c|}{-30.05\%} & \multicolumn{2}{c}{16.99\%} \\ 

\bottomrule
\end{tabular}

\end{table}
	
\paragraph{Obtained results.}
All the parameters are determined as shown in Section~\ref{subsec:parameters-beta-delta}.
Table~\ref{tab:result-all} shows the average values for the indicators, as well as the average standard deviation (in italic) obtained by each algorithm over all values of $m$.
The average standard deviation is only used to give an idea of the order of magnitude of the stability of each algorithm.
Since Simple \KMeans{}, Temporal-Driven \KMeans{} and Constrained \KMeans{} are designed to optimize mainly one component, it is not surprising that they show the best scores for, respectively, the multidimensional variance, the temporal variance and the entropy (best results in boldface). 
\TDCKMeans{} seeks to provide a compromise, obtaining in two out of three cases the second best score.
It is noteworthy that the proposed temporal-aware dissimilarity measure used in Temporal-Driven \KMeans{} provides the highest stability (the lowest average standard deviation) for all indicators.
Meanwhile, the constrained algorithms (Constrained \KMeans{} and tcK-Means) show high instability, especially on \textit{Tvar}.
\TDCKMeans{} shows a very good stability.
The second part of Table~\ref{tab:result-all} gives the relative gain of performance of each of the proposed algorithms over Simple \KMeans{}.
It is noteworthy the effectiveness of the temporal-aware dissimilarity measure proposed in Section~\ref{subsec:proposal-temporal-dissimilarity-measure}, with a 58\% gain of Temporal Variance and less than 2\% loss of multidimensional variance.
The proposed dissimilarity measure greatly enhances the temporal cohesion of the resulted clusters, without a significant scattering of observations in the multidimensional space.
Similarly, the Constrained KM shows an improvement in the contiguity measure \textit{ShaP} of 42\%, while losing 10\% multidimensional variance.
By comparison, tcK-Means shows modest results, improving \textit{ShaP} by only 17\% and still showing important losses on both \textit{Tvar} (-30\%) and \textit{MDvar} (-2\%).
This proves that the threshold penalty function proposed in literature has lower performances than our newly proposed contiguity penalty function.
\TDCKMeans{} combines the advantages of the other two algorithms, providing an important gain of 43\% of temporal variance and increasing the \textit{ShaP} measure by more than 4\%.
Nonetheless, it shows a 6\% loss of \textit{MDvar}.

\paragraph{Varying the number of clusters}
Similar conclusions can be drawn when varying the number of clusters.
\textit{MDvar} (Figure~\ref{subfig:mdvar-noclus}) decreases, for all algorithms, as the number of cluster increases.
It is well known in clustering literature that the intra-cluster variance decreases steadily with the increase of number of clusters.
As the number of clusters augments, so does the differences of \TDCKMeans{} and Constrained \KMeans{}, when compared to the Simple \KMeans{} algorithm.
This is due to the fact that the constraints do not let too many clusters to be assigned to the same entity, resulting in the convergence towards a local optimum, with a higher value of \textit{MDvar}.
An opposite behavior is shown by the \textit{ShaP} measure in Figure~\ref{subfig:shap-noclus}, which increases with the number of clusters.
It is interesting to observe how the \textit{MDvar} and the \textit{ShaP} measures have almost opposite behaviors.
An algorithm that shows the best performances on one of the measures, also shows the worst on the other.
The temporal divergence in Figure~\ref{subfig:tvar-noclus} shows a very sharp decrease for a low number of clusters, and afterwards remains relatively constant.

\subsection{Impact of parameters $\beta$ and $\delta$}
\label{subsec:parameters-beta-delta}

\begin{figure}[!t]
	\centering

	\subfloat[]{
		\includegraphics[width=0.45\textwidth]{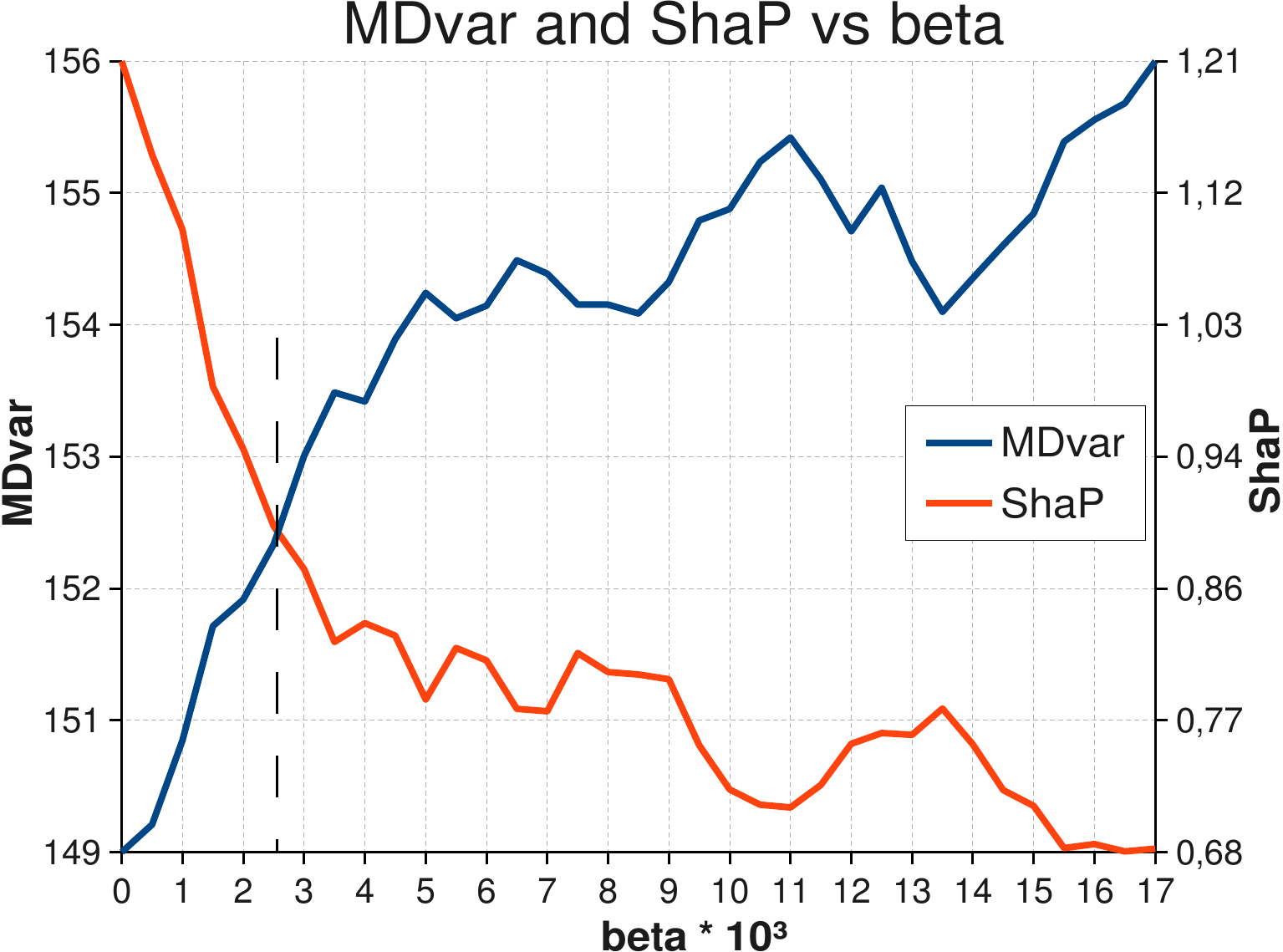}%
		\label{subfig:mdvar-shap-beta}
	}
	\hfill
	\subfloat[]{
		\includegraphics[width=0.45\textwidth]{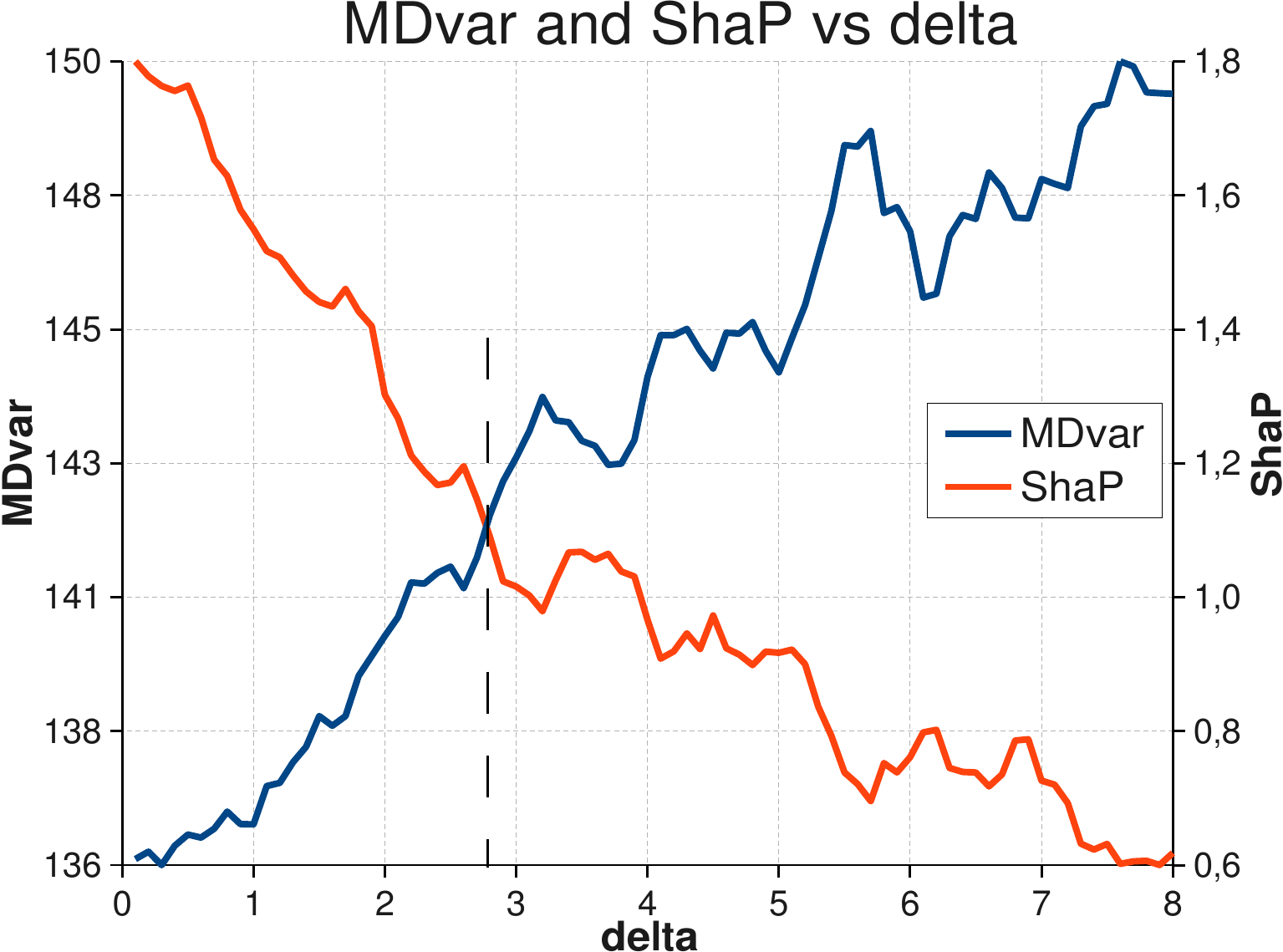}%
		\label{subfig:mdvar-shap-delta}
	}
	
	\caption{\textit{MDvar} and \textit{ShaP} function of $\beta$ (a) and of $\delta$ (b)}
	\vspace{-0.2in}
\end{figure}

The $\beta$ parameter controls the impact of the contiguity constraints in Equation~(\ref{eq:penalty-function}).
When set to zero, no constraints are imposed, and the algorithm behaves just like the Simple \KMeans{}.
The higher the values of $\beta$, the higher the penalty inflicted when breaking a constraint.
When $\beta$ is set to large values, the penalty factor will take precedence over the similarity measure in the objective function.
Observations that belong to a certain entity will be assigned to the same cluster, regardless of their resemblance in the description space.
When this happens, the algorithm cannot create partitions with higher number of clusters than the number of entities.
In order to evaluate the influence of parameter $\beta$, we execute the Constrained \KMeans{} algorithm with $\beta$ varying from 0 to 0.017 with a step of 0.0005.
The value of $\delta$ is set at 3, and 5 clusters are constructed.
For each value of $\beta$, we executed 10 times the algorithm and we plot the average obtained values.
Figure~\ref{subfig:mdvar-shap-beta} shows the evolution of measures \textit{MDvar} and \textit{ShaP} with $\beta$.
When $\beta=0$ both \textit{MDvar} and \textit{ShaP} have the same values as for Simple \KMeans{}.
As $\beta$ increases, so does the penalty for non-contiguous segmentation of entities.
\textit{MDvar} starts to increase rapidly, while \textit{ShaP} decreases rapidly.
Once $\beta$ reaches higher values, the measures continue their evolution, but with a leaner slope.
In the extreme case, in which all observations are assigned to the same cluster regardless of their similarity, the \textit{ShaP} measure will reach zero.

The $\delta$ parameter controls the width of the penalty function in Equation~(\ref{eq:penalty-function}).
As Figure~\ref{fig:penalty-delta} shows, when $\delta$ has a low value, a penalty is inflicted only if the time difference of a pair of observations is small.
As the time difference increases, the function quickly converges to zero.
As $\delta$ increases, the function decreases with a leaner slope, thus also taking into account observations which are farther away in time.
In order to analyze the behavior of the penalty function when varying $\delta$, we have executed the Constrained \KMeans{}, with $\delta$ ranging from 0.1 to 8, using a step of 0.1.
$\beta$ was set at 0.003 and 10 clusters were constructed.
Figure~\ref{subfig:mdvar-shap-delta} plots the evolution of measures \textit{MDvar} and \textit{ShaP} with $\delta$.
The contiguity measure \textit{ShaP} decreases almost linearly as $\delta$ increases, as the series of observations belonging to each entity gets segmented in larger chunks.
At the same time, the multidimensional variance \textit{MDvar} increases linearly with $\delta$.
Clusters become more heterogeneous and variance increases, as observations get assigned to clusters based on their membership to an entity, rather than their descriptive similarities.

Varying $\alpha^*$ and $d^*$ for the tcK-Means proposed in~\cite{LIN06} yields similar results, with the \textit{MDvar} augmenting and the \textit{ShaP} descending, when $\alpha^*$ and $d^*$ increase.
For the tcK-Means, these evolutions are linear, whereas for the Constrained \KMeans{} they are exponential, following a trend line of function $e^{-\frac{const}{x}}$.
Plotting the evolution of the \textit{MDvar} and the \textit{ShaP} indicators on the same graphic, provides a heuristic for choosing the optimum values for the $(\beta, \delta)$ parameters of the Constrained \KMeans{} and the \TDCKMeans{}, respectively the $(\alpha^*, d^*)$ parameters of the tcK-Means.
Both curves are plotted with the vertical axis scaled to the interval $[min_{value}, max_{value}]$. 
Their point of intersection determines the values of the parameters (as shown in Figure~\ref{subfig:mdvar-shap-beta} and~\ref{subfig:mdvar-shap-delta}).
The disadvantage of such a heuristic would be that a large number of executions must be performed with multiple values for the parameters before the ``optimum'' can be found.

\subsection{The tuning parameter $\alpha$}
\label{subsec:parameters-alpha}

The parameter $\alpha$, proposed in Section~\ref{subsec:measure-tuning-alpha}, allows the fine tuning of the ratio between the multidimensional component and the temporal component in the temporal-aware dissimilarity measure.
When $\alpha$ is close to -1, the temporal component is predominant.
Conversely, when $\alpha$ is close to 1, the multidimensional component takes precedence.
The two components have equal weights when $\alpha = 0$.
To evaluate the influence of parameter $\alpha$, we executed Temporal-Driven \KMeans{} with $\alpha$ varying from -1 to 1 with a step of 0.1.
In order not to bias the results and to evaluate only the impact of the tuning parameter, we remove the contiguity constraints from the objective function $\mathcal{J}$, by setting $\beta = 0$.
For each value of $\alpha$, we executed 10 times and we present the average values.

\begin{figure}[!t]
	\centering

	\subfloat[]{
		\includegraphics[width=0.45\textwidth]{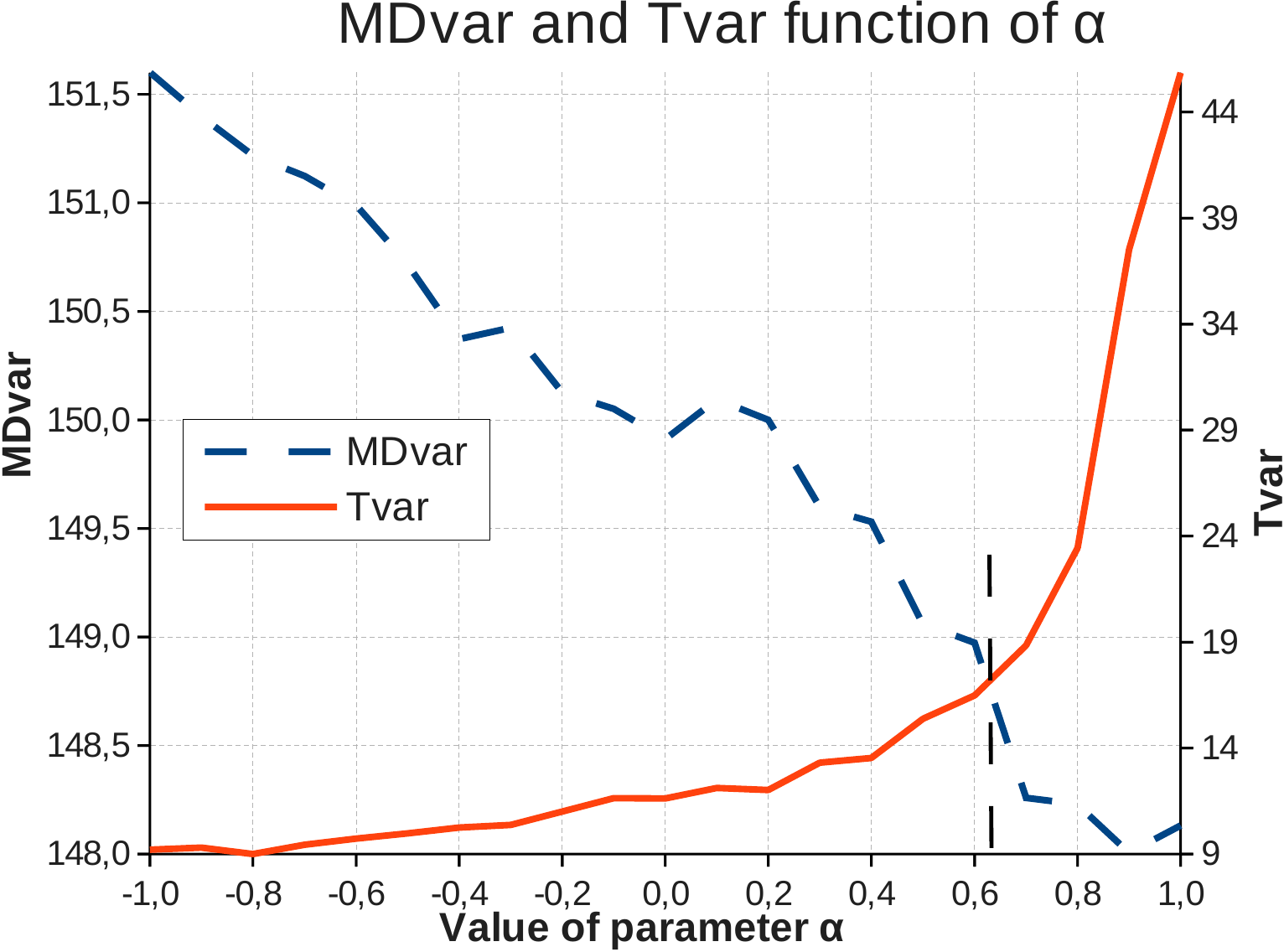}%
		\label{subfig:mdvar-tvar-alpha}
	}
	\hfill
	\subfloat[]{
		\includegraphics[width=0.45\textwidth]{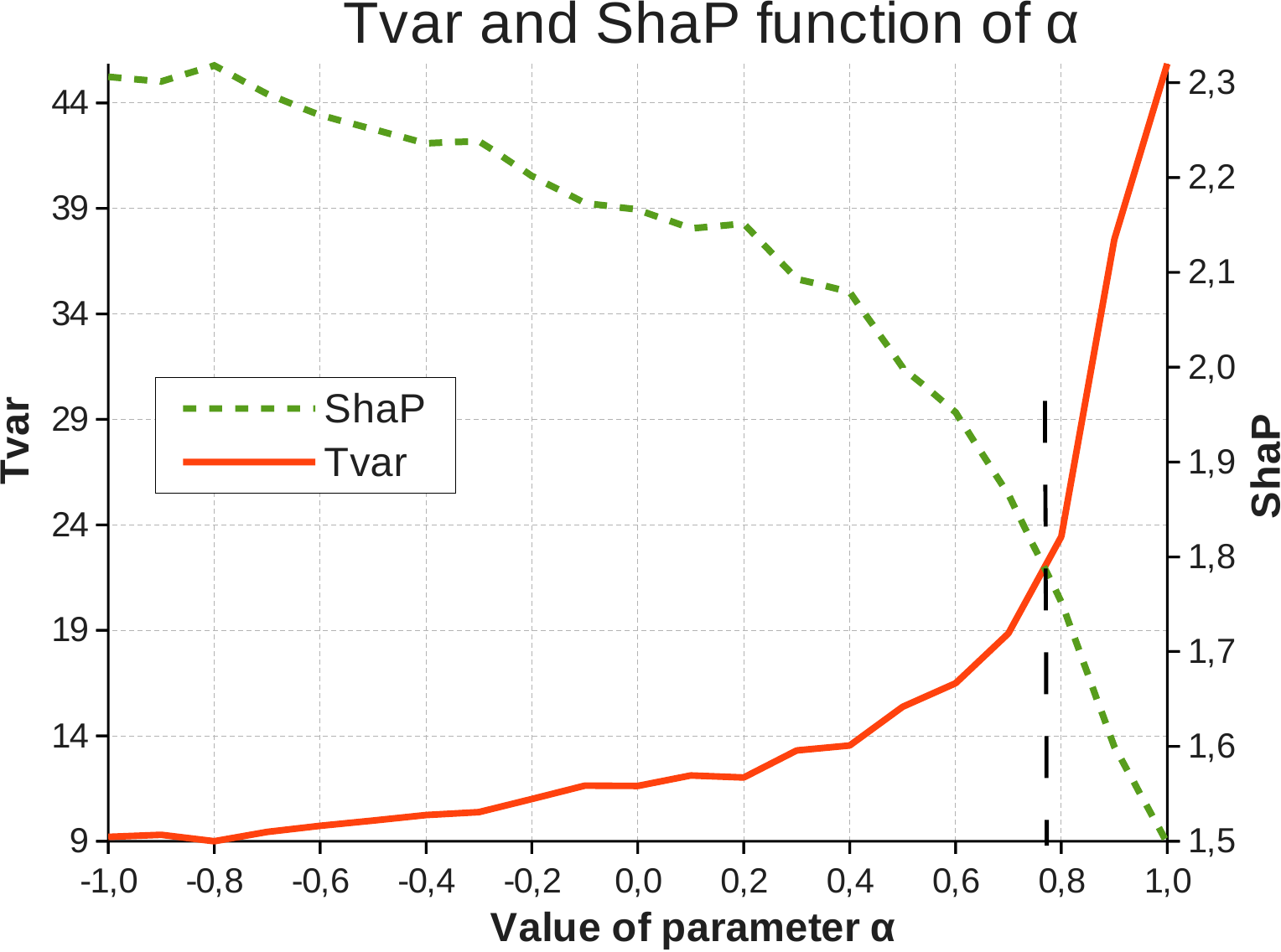}%
		\label{subfig:tvar-shap-alpha}
	}
	
	\caption{Influence of tuning parameter $\alpha$ on \textit{MDvar} and \textit{Tvar} (a) and \textit{Tvar} and \textit{ShaP} (b)}
	\label{fig:parameter-alpha}
	\vspace{-0.2in}
\end{figure}

Figure~\ref{subfig:mdvar-tvar-alpha} shows the evolution of measures \textit{MDvar} and \textit{Tvar} with $\alpha$.
For low values of $\alpha$, the value of the temporal-aware dissimilarity measure is given mainly by the temporal component, so \textit{Tvar} shows its lowest value, while \textit{MDvar} presents its maximum.
As $\alpha$ increases, \textit{MDvar} decreases as more importance is given to the multidimensional component.
For $\alpha \in (-1, 0]$, the importance of the temporal component remains intact, the increase of \textit{Tvar} is solely the result of the algorithm converging to a local optimum which also takes into account the multidimensional component.
For $\alpha \in [0, 1)$, the impact of the multidimensional component stays constant, whereas the importance of the temporal components diminishes.
As a result, \textit{MDvar} continues its decrease and \textit{Tvar} increases sharply.
For $\alpha = 1$ the temporal component is basically ignored from the measure.
The Temporal-Driven \KMeans{} behaves just like Simple \KMeans{}.
Figure~\ref{subfig:tvar-shap-alpha} shows the evolution of \textit{ShaP} alongside \textit{MDvar}.
Even if the contiguity penalty component was neutralized by setting $\beta = 0$, the value of \textit{ShaP} is not constant, but it descends with $\alpha$.
For low values of $\alpha$, the temporal component is predominant in the similarity measure.
This generates partitions where every cluster regroups all observations from a specific period, regardless of their multidimensional description.
This means that all entities have segments in all the clusters, which leads to a high value of \textit{ShaP}.

It is noteworthy that the evolution of the indicators is not linear with $\alpha$.
As $\alpha$ increases, \textit{Tvar} augments only very slowly and picks up the pace only for large values of $\alpha$.
This indicates that the temporal component has an inherent advantage over the multidimensional one.
As we presumed in Section~\ref{subsec:measure-tuning-alpha}, this is due to the intrinsic nature of the dataset and the main reason why the tuning parameter $\alpha$ was introduced.
The distributions of observations in the multidimensional and temporal spaces is different: in the temporal space, the observations tend to be evenly distributed, whereas in the multidimensional description space, they cluster together.
To quantify this, we calculate the ratio between average standard deviation and average distances between pairs of observations:
\vspace{-0.1in}
\begin{equation*}
	r_{dim} = \dfrac{1}{|\mathcal{X}|}\sum_{i=1}^{|\mathcal{X}| } \dfrac{stdev \left( \left\lbrace || x_i^{dim} - x_j^{dim} ||^2 \middle| x_j \in \mathcal{X}, \; i \neq j \right\rbrace \right)}{\dfrac{1}{|\mathcal{X}|}\sum_{\substack{j=1\\j \neq i}}^{|\mathcal{X}|} || x_i^{dim} - x_j^{dim} ||^2 }
\end{equation*}
where $dim$ is replaced with $d$ or $t$ ( the descriptive or the temporal dimension).
On \textit{Comparative Political Data Set I}, $r_{d} =  29.5\%$ and $r_{t} = 65.3\%$.
This shows that observations are a lot more dispersed in the temporal space than in the multidimensional description space.
This explains why \textit{Tvar} augments very slowly with $\alpha$ and starts to increase more rapidly only starting from $\alpha=0.4$.

Following the heuristic proposed in Section~\ref{subsec:parameters-beta-delta}, we can determine a trade-off value for $\alpha$.
As shown in Figure~\ref{fig:parameter-alpha}, all vertical axes are magnified between their functions' minimum and maximum values.
The trade-off value for $\alpha$ is found at the intersection point of \textit{MDvar} and \textit{Tvar} (and \textit{MDvar} and \textit{Shap}).
This value is set around $0.7$, showing the dataset's bias towards the temporal component.
This technique for setting the value of the tunning parameter is just a heuristic, the actual value of $\alpha$ is dependent on the dataset.
This is why we are currently working on a method, inspired from multi-objective optimization using evolutionary algorithms~\cite{ZHA07} to automatically determine the values of $\alpha$, as well as the other parameters of \TDCKMeans{} ($\beta$, $\delta$ and $\gamma$).

\section{Conclusion and future work}
\label{sec:conclusion}

In this paper, we have studied the detection of typical evolutions from a collection of observations.
We have proposed a novel way to introduce temporal information directly into the dissimilarity measure, weighting the Euclidean component in the description space by the temporal component.
We have proposed \TDCKMeans{}, an extension of \KMeans{}, which uses the temporal-aware dissimilarity measure and a new objective function which takes into consideration the temporal dimension.
We use a penalty factor to make sure that the observation series related to an entity get segmented into continuous chunks.
We infer a new centroid update formula, where elements distant in time contribute less to the centroid than the temporally close ones.
We have proposed an intersection similarity measure between two temporal clusters and a method to calculate \textit{a posteriori} an adjacency matrix.
We use this adjacency matrix in order to structure the detected evolution phases as a graph.
From a qualitative point of view, we have shown that our algorithm is capable of detecting comprehensible evolutions based on a Political Science dataset.
Quantitatively, we have shown that our proposition consistently improves temporal variance, without any significant losses in the multidimensional variance.

We are currently experimenting with applying the algorithm to other applications, \textit{e.g.}, detection of social roles in social networks, by passing through temporal behavioral roles.
A social role is defined as a typical succession of behavioral roles.
In our current work, we have inferred a temporal cluster graph structure \textit{a posteriori} to the construction of the clusters.
We have ongoing work toward incorporating the graph construction into the \TDCKMeans{} algorithm, by modifying the objective function in order to take into account the intersection similarity measure and a temporal distance.
Another direction of research will be describing the clusters in a human readable form.
We work on means to provide them with an easily comprehensible description by introducing temporal information into an unsupervised feature construction algorithm.
We are also experimenting a method for setting automatically the values of \TDCKMeans{}'s parameters ($\alpha$, $\beta$, $\delta$ and $\gamma$), by using an approach inspired from multi-objective optimization using evolutionary algorithms~\cite{ZHA07}.

{\scriptsize
\bibliographystyle{plain}
\bibliography{biblio-tempered}
}

\end{document}